\documentclass{article}

\PassOptionsToPackage{numbers, compress}{natbib}
    \usepackage[preprint]{neurips_2025}

\usepackage[utf8]{inputenc} 
\usepackage[T1]{fontenc}    
\usepackage{hyperref}       
\usepackage{url}            
\usepackage{booktabs}       
\usepackage{amsfonts}       
\usepackage{nicefrac}       
\usepackage{microtype}      
\usepackage{xcolor}         

\usepackage{amsmath}
\usepackage{graphicx}       
\usepackage{algorithm}
\usepackage{algpseudocode}
\usepackage{subcaption}      
\usepackage{todonotes}
\usepackage{booktabs}
\usepackage{multirow}
\usepackage{siunitx}
\usepackage{geometry}
\usepackage{tikz}
\usetikzlibrary{calc,fit}
\usepackage{wrapfig}

\title{dreaMLearning: Data Compression Assisted Machine Learning}

\author{%
  Xiaobo Zhao \\
  Aarhus University\\
  \texttt{xiaobo.zhao@ece.au.dk} \\
  \And
  Aaron Hurst \\
  Aarhus University \\
  \texttt{ah@ece.au.dk} \\
  \AND
  Panagiotis Karras \\
  University of Copenhagen \& Aarhus University \\
  \texttt{piekarras@gmail.com} \\
  \And
  Daniel E. Lucani \\
  Aarhus University \\
  \texttt{daniel.lucani@ece.au.dk} \\
}

\begin{document}
\maketitle

\begin{abstract}
Despite rapid advancements, machine learning, particularly deep learning, is hindered by the need for large amounts of labeled data to learn meaningful patterns without overfitting and immense demands for computation and storage, which motivate research into architectures that can achieve good performance with fewer resources. This paper introduces dreaMLearning, a novel framework that enables learning from compressed data without decompression, built upon Entropy-based Generalized Deduplication (EntroGeDe), an entropy-driven lossless compression method that consolidates information into a compact set of representative samples. DreaMLearning accommodates a wide range of data types, tasks, and model architectures. Extensive experiments on regression and classification tasks with tabular and image data demonstrate that dreaMLearning accelerates training by up to~$8.8\times$, reduces memory usage by~$10\times$, and cuts storage by~$42\%$, with a minimal impact on model performance. These advancements enhance diverse ML applications, including distributed and federated learning, and tinyML on resource-constrained edge devices, unlocking new possibilities for efficient and scalable learning.
\end{abstract}

\section{Introduction}

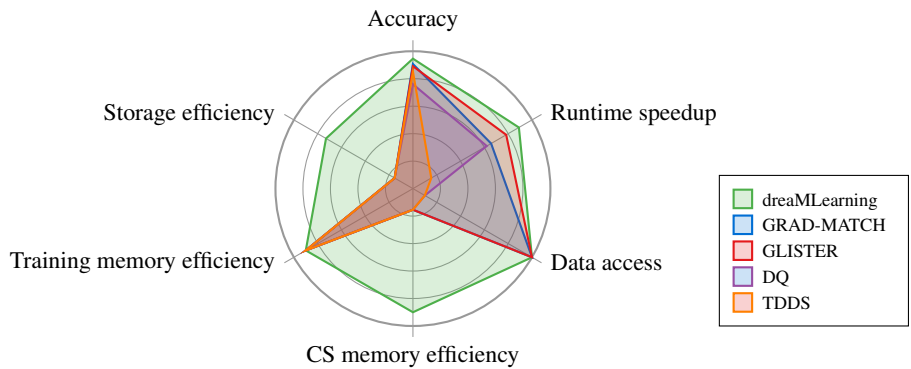
\begin{figure}[h!]
    \centering
    \newcommand{\radarradius}{5.2em}
\newcommand{\radarradiusmin}{0.8em}

\definecolor{colour1}{HTML}{006ed4}  
\definecolor{colour2}{HTML}{e41a1c}  
\definecolor{colour3}{HTML}{4daf4a}  
\definecolor{colour4}{HTML}{984ea3}  
\definecolor{colour5}{HTML}{ff7f00}
\definecolor{colour6}{HTML}{a65628}
\definecolor{colour7}{HTML}{f781bf}
\definecolor{colour8}{HTML}{999999}

\begin{tikzpicture}[
    grid/.style={black!40},
    metric label/.style={font=\small, black, anchor=west, align=center},
    legend item/.style={
        minimum height=0.7em,
        minimum width=0.7em,
        thick,
        draw,
        fill opacity=0.2,
        anchor=north,
        inner sep=0,
    }
    ]

    \draw[grid, thick] (0,0) circle (\radarradius);
    \draw[grid] (0,0) circle (0.8*\radarradius);
    \draw[grid] (0,0) circle (0.6*\radarradius);
    \draw[grid] (0,0) circle (0.4*\radarradius);
    \draw[grid] (0,0) circle (0.2*\radarradius);

    \draw[grid] (0,0) -- ( 30:1.08*\radarradius) node[metric label] {Runtime speedup};
    \draw[grid] (0,0) -- ( 90:1.08*\radarradius) node[metric label, anchor=south] {Accuracy};
    \draw[grid] (0,0) -- (150:1.08*\radarradius) node[metric label, anchor=east] {Storage efficiency};
    \draw[grid] (0,0) -- (210:1.08*\radarradius) node[metric label, anchor=east] {Training memory efficiency};
    \draw[grid] (0,0) -- (270:1.08*\radarradius) node[metric label, anchor=north] {CS memory efficiency};
    \draw[grid] (0,0) -- (330:1.08*\radarradius) node[metric label] {Data access};


    \draw[colour3, thick, fill, fill opacity=0.2]
    ( 30:0.890000*\radarradius) 
    -- ( 90:0.946*\radarradius) 
    -- (150:0.730*\radarradius) 
    -- (210:0.900*\radarradius) 
    -- (270:0.900*\radarradius) 
    -- (330:1.000*\radarradius) 
    -- cycle;

    \draw[colour1, thick, fill, fill opacity=0.2]
    ( 30:0.656700*\radarradius)  
    -- ( 90:0.907*\radarradius)  
    -- (150:   \radarradiusmin)  
    -- (210:0.900*\radarradius)  
    -- (270:   \radarradiusmin)  
    -- (330:1.000*\radarradius)  
    -- cycle;

    \draw[colour2, thick, fill, fill opacity=0.2]
    ( 30:0.783300*\radarradius)  
    -- ( 90:0.891*\radarradius)  
    -- (150:   \radarradiusmin)  
    -- (210:0.900*\radarradius)  
    -- (270:   \radarradiusmin)  
    -- (330:1.000*\radarradius)  
    -- cycle;

    \draw[colour4, thick, fill, fill opacity=0.2]
    ( 30:0.620000*\radarradius)  
    -- ( 90:0.760*\radarradius)  
    -- (150:   \radarradiusmin)  
    -- (210:0.900*\radarradius)  
    -- (270:   \radarradiusmin)  
    -- (330:0.100*\radarradius)  
    -- cycle;

    \draw[colour5, thick, fill, fill opacity=0.2]
    ( 30:      \radarradiusmin)  
    -- ( 90:0.851*\radarradius)  
    -- (150:   \radarradiusmin)  
    -- (210:0.900*\radarradius)  
    -- (270:   \radarradiusmin)  
    -- (330:0.100*\radarradius)  
    -- cycle;


    \node[legend item, colour3, fill=colour3] (legend1) at (0:2.4*\radarradius) {};
    \node[legend item, colour1, fill=colour1] (legend2) at ([yshift=-0.2em]legend1.south) {};
    \node[legend item, colour2, fill=colour2] (legend3) at ([yshift=-0.2em]legend2.south) {};
    \node[legend item, colour4, fill=colour1] (legend4) at ([yshift=-0.2em]legend3.south) {};
    \node[legend item, colour5, fill=colour2] (legend5) at ([yshift=-0.2em]legend4.south) {};

    \node[font=\scriptsize, anchor=west, text depth=0.1em, text height=0.6em] (label1) at (legend1.east) {dreaMLearning};
    \node[font=\scriptsize, anchor=west, text depth=0.1em, text height=0.6em] (label2) at (legend2.east) {GRAD-MATCH};
    \node[font=\scriptsize, anchor=west, text depth=0.1em, text height=0.6em] (label3) at (legend3.east) {GLISTER};
    \node[font=\scriptsize, anchor=west, text depth=0.1em, text height=0.6em] (label3) at (legend4.east) {DQ};
    \node[font=\scriptsize, anchor=west, text depth=0.1em, text height=0.6em] (label3) at (legend5.east) {TDDS};

    \node[fit=(legend1)(label2.east)(legend5), draw, black, inner sep=0.5em] () {};

\end{tikzpicture}
    \caption{
        dreaMLearning performance benefits over existing coreset selection (CS) methods.
    }
    \label{fig:performance_radar}
\end{figure}

Scaling machine learning (ML) systems to handle larger datasets and more complex model architectures places increasing demands on computational system resources, e.g., storage, memory, and processing power~\cite{shen2024efficient, menghani2023efficient,zhou2022dataset,nguyen2021dataset}.
Training high-performance models typically requires extensive and diverse datasets, but storing and accessing such data repeatedly is costly, especially in resource-constrained settings such as edge devices, federated learning.

Methods such as dataset distillation~\cite{wang2018dataset, cazenavette2022dataset} and coreset selection~\cite{sener2018active,sinha2020small,colemanselection,tonevaempirical,paul2021deep,mirzasoleiman2020coresets,killamsetty2021grad,killamsetty2021glister,zhang2024spanning} \textbf{trade-off accuracy for training speed-ups} by carrying out the training on a smaller number of samples, e.g., a carefully chosen subset of the data for coreset selection or a synthetically crafted dataset in distillation. 
However, they often rely on computationally intensive optimization and assume full dataset access during both the selection/distillation and, in many cases, the training process.
These added computational costs also affect the overall speed-up possible with these techniques. Finally, these methods tend to be tightly coupled to the model and the task.

Classical lossless data compression can be used to reduce the footprint of the dataset when stored, but require decompression and loading of the uncompressed dataset into memory for the training process, including coreset selection or dataset distillation, if applied. Lossy compression methods are available, but \textbf{trade-off storage space for accuracy} and may generate unintended results~\cite{underwood2024understanding}. In the case of images, the accuracy-compression trade-off has been shown to be non-linear, dependent on the ML task, model and data~\cite{zhao2020improving}.

This raises a fundamental question: \textit{can we develop a task- and model-independent ML system that can reduce storage costs, provide high memory efficiency during the data selection/distillation process, maintain high accuracy, and result in an overall runtime speed-up during training?}  

\begin{figure}[t]
    \centering
    \includegraphics[width=0.9\textwidth]{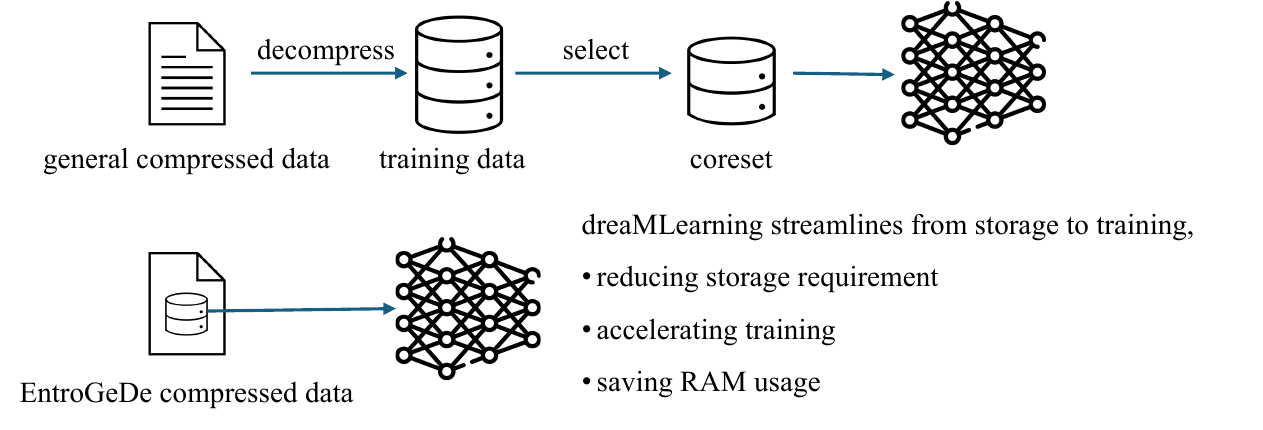}
    \caption{A comparison between the existing pipeline and the proposed dreaMLearning framework.}
    \label{fig:dreamlearning_overview}
\end{figure}

\begin{figure}[h!]
    \centering
    \begin{subfigure}[b]{0.48\textwidth}
        \includegraphics[width=0.9\textwidth]{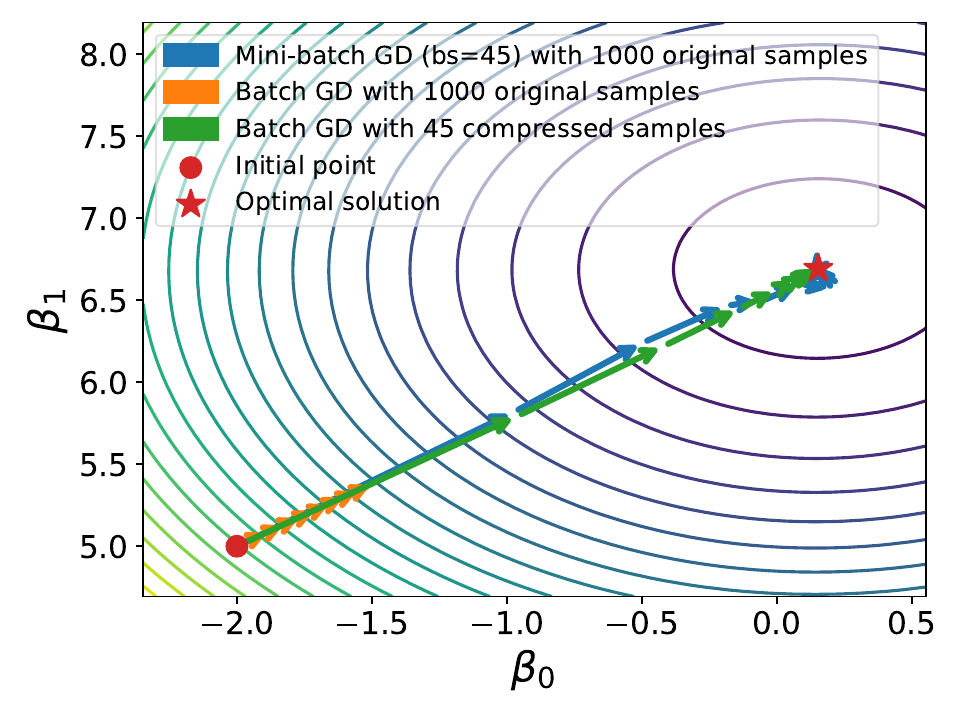}
        \caption{Convergence contour of GD.}
        \label{fig:contour}
    \end{subfigure}
    \hfill
    \begin{subfigure}[b]{0.48\textwidth}
        \includegraphics[width=0.9\textwidth]{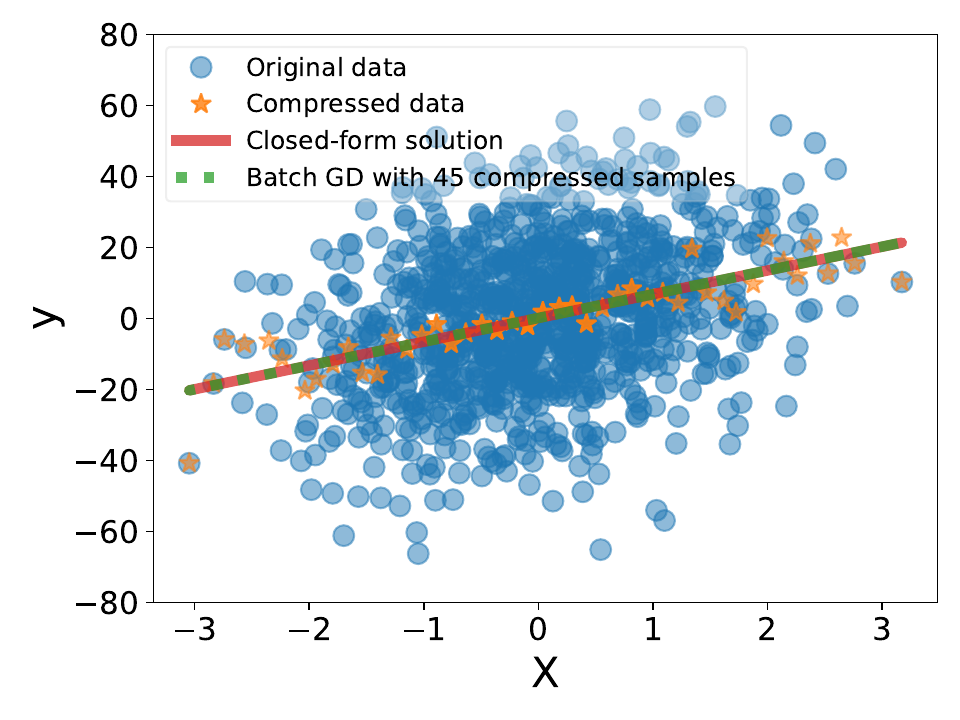}
        \caption{Compressed and original data points.}
        \label{fig:points}
    \end{subfigure}
    \caption{Linear regression with gradient descent (GD).}
    \label{fig:visualization_compressed_ML}
\end{figure}

We propose \textit{dreaMLearning}, a general-purpose framework that enables direct training on compressed data without decompression, streamlining from storage to training Fig.~\ref{fig:performance_radar} summarizes its benefits based on our experimental results.  
The framework is built upon \textit{Entropy-based Generalized Deduplication} (EntroGeDe), a lossless, random-access compression method that clusters similar data points and encodes them into condensed, representative samples guided by entropy (See Fig.~\ref{fig:dreamlearning_overview}). 
Unlike conventional pipelines that first decompress data and select coresets through computationally expensive optimization, \textit{dreaMLearning} produces training-ready compressed datasets that preserve essential learning characteristics.
Figure~\ref{fig:contour} illustrates a simple linear regression task using gradient descent, a fundamental machine learning algorithm, with uniform computational cost per step.
Training on compressed data converges more rapidly than for a batch GD with the entire dataset. It also follows the gradient more accurately than stochastic gradient descent (mini-batch GD), avoiding the oscillations of the latter and resulting in faster convergence.
Figure~\ref{fig:points} illustrates that compressed data points effectively capture the underlying data distribution, including outliers, preserving essential characteristics for robust learning.

This paper shows that the \textit{dreaMLearning} framework can be implemented with different degrees of complexity and adapt to required data characteristics and performance demands. For example, simpler data sets and problems can rely on a fixed selection of samples, while more complex data sets may introduce lightweight random sampling approaches, and even frequency-domain transformations to improve compressibility for storage in some cases while maintaining memory and accuracy gains.
This flexible design allows \textit{dreaMLearning} to support diverse data modalities, model architectures, and learning tasks.

This work makes the following contributions.  
First, we introduce dreaMLearning, a unified framework that enables training directly on compressed data, eliminating the need for decompression or subset selection.  
Second, we propose EntroGeDe, an entropy-based extension of Generalized Deduplication that jointly optimizes information retention and compression efficiency, two traditionally conflicting objectives in prior GeDe methods.  
Third, through extensive experiments on tabular and image datasets, we show that dreaMLearning significantly improves training speed, memory usage, and storage efficiency, while maintaining competitive accuracy.

\section{Background and related work}


This section reviews foundational techniques that underpin our approach, focusing on two key areas: data compression via Generalized Deduplication (GeDe) and coreset selection for efficient model training.  
We first describe GeDe, which enables lossless compression with efficient random access, and discuss recent extensions to various data types and analytical tasks.  
Next, we survey coreset selection methods, and highlight their limitations in scalability and computational overhead.  
Finally, we introduce our proposed method, dreaMLearning, which unifies these two techniques to enable efficient model training directly on compressed data.

\subsection{Generalized deduplication}

\begin{figure}[h!]
    \centering
    \includegraphics[width=0.42\linewidth]{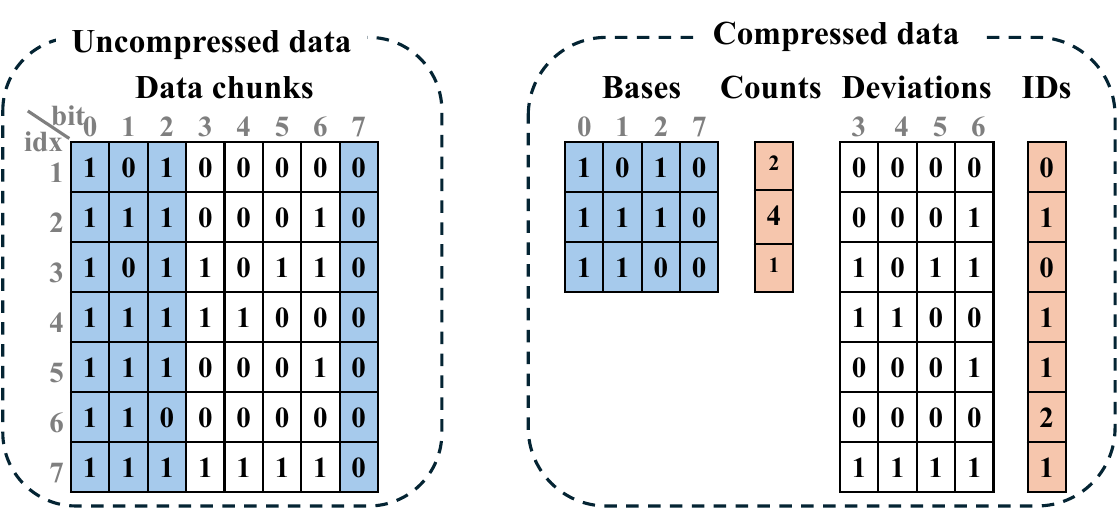}
    \caption{An example of GeDe applied to 8-bit data chunks.}
    \label{fig:GD}
\end{figure}


Deduplication~\cite{quinlan2002venti, meyer2012study} compresses data by replacing identical data chunks with pointers to their first occurrence. Generalized Deduplication~(GeDe)~\cite{Vestergaard_2019b} extends this technique to encompass similar, albeit non-identical data chunks. As illustrated in Figure~\ref{fig:GD}, GeDe splits data chunks into frequently appearing parts, \emph{bases}, and high-variance parts, \emph{deviations}, and deduplicates bases and stores deviations unchanged alongside pointers to corresponding bases to enable lossless decompression. Unlike many other lossless compression algorithms, GeDe also supports efficient random access~\cite{Vestergaard_2020}.

The key question that determines GeDe compression performance is how and where to split data chunks into bases and deviations. In general, allocating more data to the \emph{base} (i.e., the collections of all bases) improves compression since more data is deduplicated, but can also increase the number of unique bases, which reduces compression~\cite{hurst2022glean}. Various heuristic methods have been proposed for allocating data between base and deviation. One early approach pre-computes inter-bit correlations and uses the maximum correlation for each bit to select bits for allocation to the base~\cite{Vestergaard_2020}.

A recent variant, GreedyGD~\cite{hurst2024greedygd}, uses a greedy search algorithm to iteratively minimize the number of new bases created by enlarging the base. GreedyGD also introduces data pre-processing that significantly improves compression using GeDe; it has been further refined for floating-point data~\cite{Taurone_2023}. GeDe has also been applied to image compression, where it performs particularly well~\cite{Rask_2024}.

As it enables efficient random access, GeDe also facilitates analytics directly on compressed data without decompression~\cite{Hurst_2021}. Moreover, accessing only the bases elements suffices for approximate analytics, accelerating tasks. For instance, one may perform highly accurate $k$-means clustering on GeDe-compressed data much faster and with lower memory usage than with uncompressed data~\cite{Hurst_2021,hurst2022glean}. Similar results have also been achieved for anomaly detection~\cite{Taurone_2024}, while allocating additional bits to the base improves performance~\cite{hurst2022glean}.
\subsection{Coreset selection}

Coreset selection (CS) identifies subsets of data that retain the essential learning characteristics of the full dataset. 
Several strategies have been proposed, including geometry-based methods~\cite{sener2018active, sinha2020small}, uncertainty-based methods~\cite{colemanselection}, and error/loss-based methods~\cite{tonevaempirical,paul2021deep}. 
The state-of-the-art rests on gradient-based methods~\cite{mirzasoleiman2020coresets, killamsetty2021grad, killamsetty2021glister}, which leverage gradients computed during training to select data points aligned with the model's optimization dynamics. 
CRAIG~\cite{mirzasoleiman2020coresets} and GRAD-MATCH~\cite{killamsetty2021grad} iteratively select subsets whose gradients match those of the full data, leading to comparable training dynamics. 
GLISTER~\cite{killamsetty2021glister} formulates subset selection as a bi-level optimization problem, maximizing validation performance rather than minimizing training loss. 
Still, these approaches are computationally intensive, as they require multiple rounds of gradient computation and optimization, and become particularly burdensome on large-scale datasets or complex models with costly gradient evaluations.

To address this predicament, another line of work uses static subsets during training to avoid updates. 
TDDS~\cite{zhang2024spanning} ranks samples based on their contribution to training, combining temporal consistency with gradient-based metrics. 
Nevertheless, TDDS requires training on the full data upfront. 
Dataset Quantization (DQ)~\cite{zhou2023dataset} clusters data points and forms a coreset by representatives from each cluster guided by submodular maximization~\cite{iyer2021submodular} to effectively reduce data size while preserving essential information. 
Nevertheless, a fixed subset may not always optimize model performance.

Contrariwise to these methods, dreaMLearning integrates lossless data compression into coreset construction into a single pipeline to enable scalable and effective model training from compact data representations. 
It employs \emph{entropy-based generalized deduplication} (EntroGeDe) to extract a compact set of aggregate representations from clusters of similar data points, which form \emph{condensed} rather than raw samples, thereby capturing rich statistics. 
The extracted coreset is \emph{training-ready} without decompression or further selection. 
Both selection and compression are \emph{entropy-driven}, offering a trade-off between compression power and learning utility.

\section{Entropy based generalized deduplication}

In existing GeDe-based compression methods~\cite{Vestergaard_2020,hurst2022glean,hurst2024greedygd,Taurone_2023,Rask_2024,Hurst_2021,Taurone_2024}, the base can be accessed directly without decompression. Combined with some (small) meta-data generated during the compression process, e.g., the number of times each base is used, these bases can serve as an interesting summary of the original data that support  approximate, yet very accurate analytics, e.g.,~\cite{hurst2022glean, hurst2024greedygd}.
Selecting base bits for GeDe typically balances two conflicting objectives: compression efficiency and analytics utility.
If the selected bits that form the bases generate a lot of duplicates of the bases for each sample, i.e., few unique bases compared to the number of samples in the dataset, this results in a high compression. However, a given bit selection also affects the quality of meaningful analytics in two ways. First, a larger number of bases results in a richer summary for the data, while potentially reducing compression efficiency. Second, if bits from say a given dimension corresponds to the most significant bits for the value of the data, then the analytic calculation is likely to be more accurate compared to choosing the least significant bits.  
For analytics calculations, existing methods assume deviation bits to be zero~\citep{Hurst_2021,hurst2022glean,hurst2024greedygd}.

In contrast to past methods, \textbf{EntroGeDe} leverages entropy per bit position to guide both clustering and compression processes, as outlined in Algorithm~\ref{alg:entroGeDe}.
This entropy-guided selection allows EntroGeDe to balance information retention and data deduplication. Additionally, we compute the average deviation per base to improve accuracy compared to previous zero-filling strategies.

\paragraph{Intrinsic clustering process} We preprocess the dataset using the GreedyGeDe method~\citep{Hurst_2021}, converting it to binary format and calculating the entropy of each bit position. For each of the $d$ columns, we select the most significant non-constant bits (MSBs), rank them by decreasing entropy, and choose the top $\beta$ bits, proceeding to the next MSBs if $\beta > d$ until $\beta$ is reached. Using these $\beta$ bits, we cluster the $n$ data points into $m$ clusters ($m \ll n$) based on their unique bit combinations. The centroid of each cluster serves as a condensed sample, weighted by the cluster size. These samples are appended to the original dataset for further compression, with minimal impact on the compression ratio due to $m \ll n$. In storage-constrained environments, we store only the $\beta$ bit positions, generating representative samples on demand.

\begin{algorithm}[H]
\caption{Entropy-based Generalized Deduplication (EntroGeDe)}
\label{alg:entroGeDe}
\begin{algorithmic}[1]
\Require Dataset $\mathcal{D}$ with $n$ data points; analytics bit count $\beta$; plateau threshold $\tau$
\Ensure Compressed dataset $\mathcal{D}'$ consisting of selected base bit positions $B_{\text{pos}}^*$, bases $\mathbf{B}$, deviations $\boldsymbol{\Delta}$ with base IDs, and condensed sample weights $\mathbf{w}$

\State Preprocess and convert $\mathcal{D}$ to binary format; let $l_t \gets$ total number of bits per data point
\State Compute entropy $H(p_i)$ for each bit position $p_i$ in $\mathcal{D}$

\State \textcolor{gray}{\textit{// Clustering phase (prioritizing high-entropy bits)}}
\State Consider MSBs of all $d$ columns first; if $\beta > d$, proceed to next MSBs
\State Sort bit positions in decreasing order of entropy
\State Select top $\beta$ bit positions for clustering
\State Cluster data points based on matching values at selected bit positions
\State Compute cluster centroids as condensed samples; record cluster sizes as weights $\mathbf{w}$

\State \textcolor{gray}{\textit{// Compression phase (prioritizing low-entropy bits)}}
\State Initialize base bit positions $B_{\text{pos}} \gets$ constant positions ($v$ bits), set number of bases $n_b \gets 1$, base length $l_b \gets v$, deviation length $l_d \gets l_t - l_b$, plateau counter $c \gets 0$
\State Compute initial size $S^*$ using Equation~\ref{eq:total_size}

\State Sort remaining bit positions by increasing entropy
\For{each position $p_i$ in sorted list}
    \State Add $p_i$ to $B_{\text{pos}}$
    \State Update bases $\mathbf{B}$ at $B_{\text{pos}}$
    \State Update $n_b \gets |\mathbf{B}|$, $l_b \gets l_b + 1$, $l_d \gets l_d - 1$
    \State Compute compressed size $S$
    \If{$S < S^*$}
        \State $S^* \gets S$, $B_{\text{pos}}^* \gets B_{\text{pos}}$
        \State $c \gets 0$
    \Else
        \State $c \gets c+1$
        \If{$c \geq \tau$ or $n_b = n$}
            \State \textbf{break}
        \EndIf
    \EndIf
\EndFor

\State \Return $\mathcal{D}'$: $B_{\text{pos}}^*$, $\mathbf{B}$, $\boldsymbol{\Delta}$ with base IDs, and weights $\mathbf{w}$
\end{algorithmic}
\end{algorithm}

\paragraph{Compression considerations}
To maximize duplicate patterns, we select base bits in order of increasing entropy, prioritizing low-entropy bits shared across data points.  
We initialize with constant bit positions $B_{\text{pos}}$ of size $v$, setting the number of bases $n_b = 1$, base bit length $l_b = v$, deviation bit length $l_d = l_t - l_b$, where $l_t$ is the total bit count, and plateau counter $c = 0$.  
Each deviation’s base ID uses $\lceil \log_2(n_b) \rceil$ bits, and each condensed sample’s weight uses $\lceil \log_2(n) \rceil$ bits.  
The initial best compressed size $S^*$ is:
\begin{equation}
S = n_b l_b + (n + m)(\lceil \log_2(n_b) \rceil + l_d) + m\lceil \log_2(n) \rceil + S_{\text{params}},
\label{eq:total_size}
\end{equation}
where $n_b l_b$ denotes the size of the bases.  
$(n + m)(\lceil \log_2(n_b) \rceil + l_d)$ represents the size of deviations, including deviation bits and base IDs.  
$m\lceil \log_2(n) \rceil$ accounts for the weights’ size.  
$S_{\text{params}}$, the size of compression parameters, is typically negligible.  
Non-constant bit positions are sorted by increasing entropy.  
Each position $p_i$ is added to $B_{\text{pos}}$, updating $n_b$ based with the number of unique bases at $B_{\text{pos}}$, incrementing $l_b$ by 1, and decrementing $l_d$ by 1.  
The compressed size $S$ is recomputed using Equation~\ref{eq:total_size}.  
If $S$ is less than the current best size $S^*$, we update $S^*$ and $B_{\text{pos}}$ and reset $c$ to 0.  
Otherwise, $c$ is incremented by 1.  
The process terminates when $c$ reaches the threshold $\tau$ or $n_b = n$, indicating no further compression.  
The hyperparameter $\tau$ balances compression efficiency and computational cost.  
Upon completion, the bases, base IDs, and deviations for the optimal compression are stored.

\section{Direct learning on compressed data}\label{sec:directML}

DreaMLearning operates on bit-level data across domains, independently of specific tasks or model architectures. To demonstrate this versatility, we illustrate direct learning on compressed data through with two tasks: regression on tabular data and classification on image data.

\subsection{Regression with tabular datasets}\label{sec:directML_tabular}

\subsubsection{Compression}
Given a tabular dataset $\{(\mathbf{x}_i, y_i)\}_{i=1}^n$ with features $\mathbf{x}_i \in \mathbb{R}^d$ and targets $y_i \in \mathbb{R}$, we apply the EntroGeDe algorithm (Algorithm~\ref{alg:entroGeDe}) for compression.  
This yields $m$ condensed samples $(\mathbf{x}_j^c, y_j^c)$ with weights $w_j$, where $m \ll n$. Each sample summarizes a cluster of original data points, preserving essential patterns for learning.  
Given EntroGeDe’s fine-grained random access property, these weighted samples can be directly used for model training without full decompression, offering both computational efficiency and data fidelity.

\subsubsection{Learning}
Linear regression estimates a parameter vector $\boldsymbol{\theta} \in \mathbb{R}^d$ to model the relationship between inputs and targets via $\hat{y}_i = \mathbf{x}_i^T \boldsymbol{\theta}$.  
The mean squared error (MSE) loss is defined as:
\begin{equation}
    J(\boldsymbol{\theta}) = \frac{1}{2n} \sum_{i=1}^{n} \left( \mathbf{x}_i^T \boldsymbol{\theta} - y_i \right)^2.
\end{equation}
The optimal parameters, $\boldsymbol{\theta}^*$, minimize this loss function, typically achieved through gradient descent (GD), which iteratively updates the parameters in the direction of the negative gradient, as
\begin{equation}
    \boldsymbol{\theta}_{t+1}
    = \boldsymbol{\theta}_{t} - \frac{\alpha}{n} \sum_{i=1}^{n} \nabla_{\boldsymbol{\theta}} J(\boldsymbol{\theta}_{t})
    = \boldsymbol{\theta}_{t} - \frac{\alpha}{n} \sum_{i=1}^{n} \left( \mathbf{x}_i^T \boldsymbol{\theta}_t - y_i \right) \mathbf{x}_i,
    \label{eq:linear_gd_original}
\end{equation}
where $\alpha$ is the learning rate, and $\nabla_{\boldsymbol{\theta}} J(\boldsymbol{\theta}_{t})$ is the gradient of the loss function with respect to the parameters $\boldsymbol{\theta}$ at iteration $t$.
To support learning on compressed data, we adapt this process using the $m$ weighted condensed samples from EntroGeDe.  
The weighted loss becomes:
\begin{equation}
    J_c(\boldsymbol{\theta}) = \frac{1}{2n} \sum_{j=1}^{m} w_j \left( {\mathbf{x}_j^c}^T \boldsymbol{\theta} - y_j^c \right)^2,
\end{equation}
with the corresponding update rule:
\begin{equation}
    \boldsymbol{\theta}_{t+1} = \boldsymbol{\theta}_{t} - \frac{\alpha}{n} \sum_{j=1}^{m} w_j \left( {\mathbf{x}_j^c}^T \boldsymbol{\theta}_t - y_j^c \right) \mathbf{x}_j^c.
    \label{eq:linear_gd_rep}
\end{equation}
This formulation enables efficient training directly on EntroGeDe compressed data, significantly reducing computational overhead while preserving the underlying structure and predictive power of the original data.

\subsection{Classification with compressed image datasets}
\label{sec:directML_image}

Direct learning from compressed image data presents greater challenges than regression on tabular data, primarily due to the high dimensionality and intrinsic variability of images.  
Two key issues arise: (1) condensed samples may fail to capture the dataset’s full diversity, and (2) limited inter-sample similarity hinders effective deduplication and compression.  
To address these challenges, we adopt a dual strategy: (1) randomly sampling a subset of images for training and (2) applying a frequency domain transformation to improve compression efficiency.

\subsubsection{Compression}

As shown in Section~\ref{sec:experiments}, randomly sampled subsets achieve performance comparable to state-of-the-art coreset selection methods with substantially lower computational cost.  
To leverage this efficiency, we adopt a simplified EntroGeDe approach that omits clustering and focuses exclusively on compression.  
Consequently, $m = 0$ in Equation~\ref{eq:total_size}, as no condensed samples are generated.


We adopt class-wise compression to reduce computational overhead and exploit intra-class similarities, achieving better compression ratios than cross-class approaches.  
For datasets with high redundancy, such as MNIST, direct application of EntroGeDe yields significant space savings.  
However, complex datasets like CIFAR-10 and CIFAR-100, with greater variability and fewer redundancies, are less amenable to direct compression.

To mitigate this, we apply the Discrete Cosine Transform (DCT) to shift images from the spatial to the frequency domain.  
This provides two key advantages: (1) concentration of energy in fewer coefficients, enhancing compressibility, and (2) revealing latent similarities not apparent in the spatial domain.  
The transformation introduces negligible data loss, limited to minor rounding errors with no significant impact on ML/DL performance.  
Each RGB image is first converted to the YCbCr color space (without subsampling), then DCT is applied independently to each channel.  
EntroGeDe is then used to compress the transformed data, significantly improving compression efficiency.

\subsubsection{Learning}

During training, a random subset of compressed images is retrieved directly from storage, avoiding full dataset decompression.  
This low-cost access allows subsets to be updated each epoch, maintaining exposure to diverse data throughout training.  
For DCT-compressed datasets, each retrieved image is first inverse-transformed to the spatial domain and then converted back to RGB.
Training follows standard procedures without the need of modifying the loss function or other components.

\subsection{Advantages of dreaMLearning}
The dreaMLearning framework, which integrates compression and learning, offers several advantages over traditional methods.
First, it reduces storage requirements through effective compression.  
Second, it minimizes memory usage by loading only necessary image subsets into RAM.  
Third, it accelerates training due to the smaller dataset size.  
Fourth, it eliminates the need for full dataset decompression, streamlining the pipeline.  
These advantages come with minimal impact on model performance, as demonstrated in Section~\ref{sec:experiments}.

\section{Experiments}
\label{sec:experiments}

This section evaluates dreaMLearning's performance across multiple datasets and models, benchmarking it against state-of-the-art methods. We assess accuracy, training time, storage efficiency, RAM usage for coreset selection and training, and total data access during training.

\paragraph{Baselines}

For regression tasks, dreaMLearning is compared to training on the full dataset.
For classification tasks, it is evaluated against state-of-the-art methods: GRAD-MATCH~\cite{killamsetty2021grad}, GLISTER~\cite{killamsetty2021glister}, DQ~\cite{zhou2023dataset}, TDDS~\cite{zhang2024spanning}, full dataset training, and random sampling.
Following their original protocols, GRAD-MATCH and GLISTER update subsets every 20 epochs.
DQ and TDDS use fixed subsets.
dreaMLearning and random sampling update subsets each epoch.

\paragraph{Datasets and models}

Experiments are conducted on the California Housing~\cite{pace1997sparse}, MNIST~\cite{lecun1998gradient}, CIFAR-10, and CIFAR-100~\cite{krizhevsky2009learning} datasets.
Models include linear regression with gradient descent for California Housing, LeNet5 for MNIST, and ResNet18 for CIFAR-10 and CIFAR-100.

\paragraph{Training settings}
For the California Housing dataset, 80\% of the 20,640 samples, each with 8 features and 1 target, are used for full training, with the remaining 20\% reserved for testing.
Training on compressed data utilizes condensed samples with associated weights stored within the compressed data.
A linear regression model is trained using batch gradient descent (learning rate 0.001) until convergence on an Apple M3 Pro chip with 18 GB RAM.
Results are averaged over 10 runs.


For MNIST, CIFAR-10, and CIFAR-100, standard training and testing splits are used, with a subset size of 10\% of the training set.
MNIST employs LeNet5 (100 epochs, batch size 128, SGD optimizer, cosine learning rate decay, initial learning rate 0.05, weight decay 0.0005, momentum 0.9).
CIFAR-10 and CIFAR-100 use ResNet18 (200 epochs, same hyperparameters).
Training is performed on an NVIDIA Tesla P100 GPU with 16 GB RAM, with results averaged over 5 runs.

\paragraph{Metrics}
We measure test MSE and accuracy, total time for coreset selection and training, storage requirements, RAM usage for coreset selection and training, and total data access during training.

\paragraph{Regression results}

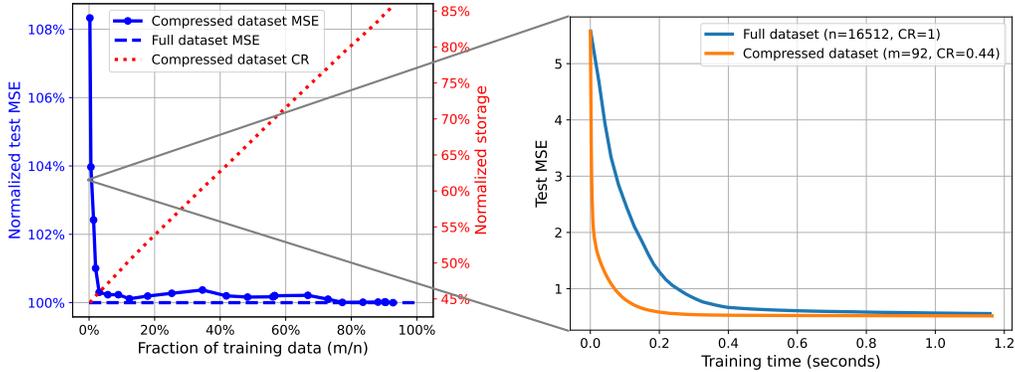
\begin{figure}[t]
\centering
\begin{tikzpicture}[remember picture]
  \node[inner sep=0pt] (subfigA) at (0,0) {%
    \begin{subfigure}[b]{0.48\textwidth}
      \centering
      \includegraphics[width=\textwidth]{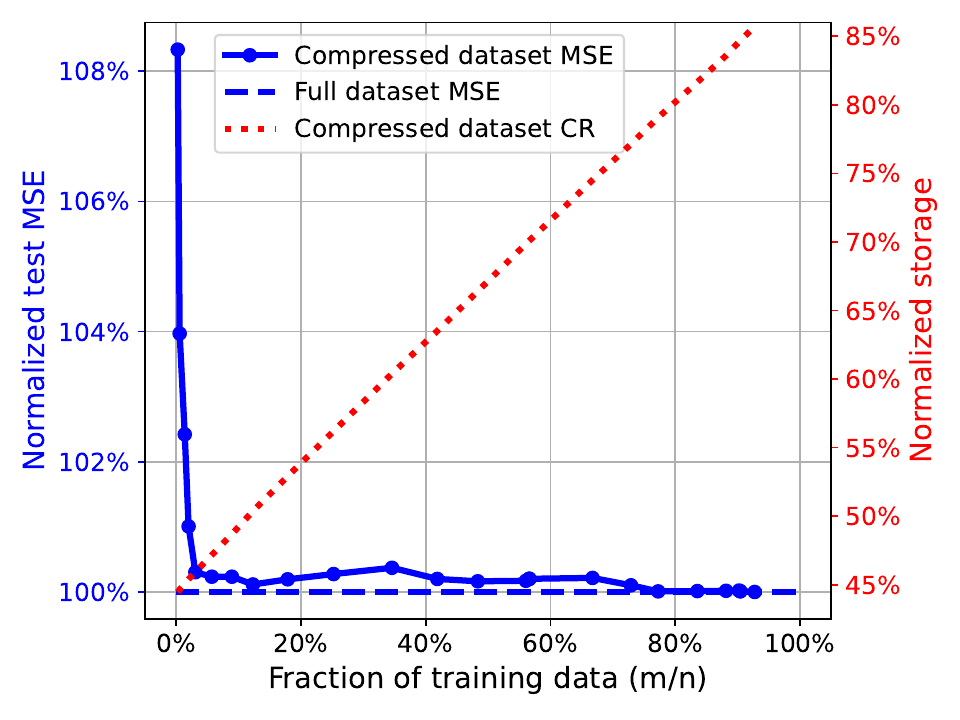}
      \caption{Compressed data fraction vs. MSE , and compressed data fraction vs. storage.}
      \label{subfig:housing_n_samples_vs_mse}
    \end{subfigure}
  };
  
  \node[inner sep=0pt] (subfigB) at (7,0) {%
    \begin{subfigure}[b]{0.48\textwidth}
      \centering
      \includegraphics[width=\textwidth]{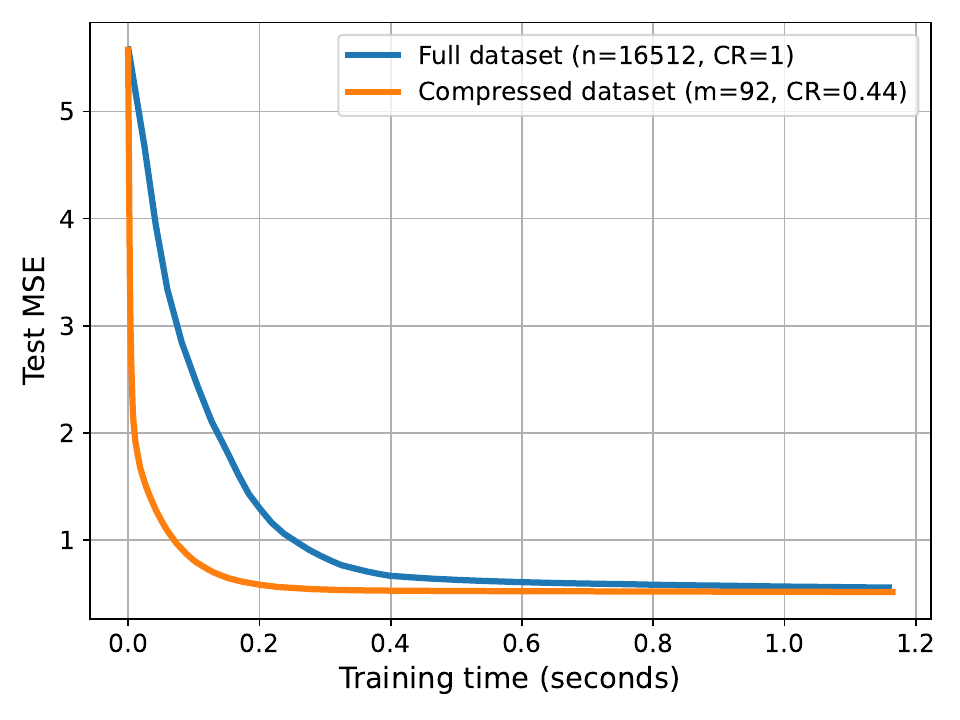}

      \caption{Training time vs. MSE.}
      \label{subfig:housing_time_vs_mse}
    \end{subfigure}
  };
  
  \coordinate (pointA) at ($(subfigA.west)!0.2!(subfigA.east) + (-0.13,0.47cm)$);

  \coordinate (pointB) at ($(subfigA.west)!1!(subfigA.east) + (0.9, 2.65cm)$);
  \coordinate (pointC) at ($(subfigA.west)!1!(subfigA.east) + (0.9, -1.54cm)$);
  
  
  \draw[gray, thick, ] (pointA) -- (pointB) node[midway, above, sloped, black] {};
  \draw[gray, thick, ] (pointA) -- (pointC) node[midway, below, sloped, black] {};
\end{tikzpicture}
\caption{Linear regression on California Housing dataset.}
\label{fig:main}
\end{figure}

Figure~\ref{subfig:housing_n_samples_vs_mse} shows the MSE of linear regression on the California Housing dataset across varying fractions of condensed data and their storage requirements.
Using condensed data equivalent to 5\% of the training set, dreaMLearning achieves performance comparable to the full dataset while requiring less than 50\% of the storage.
Figure~\ref{subfig:housing_time_vs_mse} demonstrates that dreaMLearning, with 92 condensed samples (0.6\% of training data, 44\% storage), yields an MSE only 4\% higher than the full dataset.
Its entropy-based clustering in EntroGeDe accelerates convergence by efficiently compressing information into fewer samples.



\paragraph{Classification results}
With a fixed 10\% subset size, coreset selection memory, training memory, and total data access remain consistent across MNIST, CIFAR-10, and CIFAR-100.
These metrics are thus reported only in Table~\ref{tab:mnist}.
dreaMLearning achieves competitive accuracy with significant resource savings.
On MNIST, it reaches 99.1\% accuracy (vs. 99.3\% for full dataset).
On CIFAR-10, 90.2\% (highest among subset methods).
On CIFAR-100, 69.7\% (highest among subset methods).
Storage is reduced to 58\% (MNIST), 80\% (CIFAR-10), and 73\% (CIFAR-100), with coreset selection and training memory at 10\% and training time at 11\% of the full dataset.
Compared to GRAD-MATCH and GLISTER, dreaMLearning offers comparable or higher accuracy with lower storage and memory usage.
DQ and TDDS, using fixed subsets, suffer significant accuracy losses.
Lightweight subset updates in dreaMLearning and random sampling, performed each epoch, incur no additional memory or time costs.
These results demonstrate dreaMLearning's robust balance of accuracy and efficiency across datasets of varying complexity, consistently outperforming or matching baselines.

\begin{table}[h]
\centering
\caption{MNIST}
\label{tab:mnist}
\begin{tabular}{lcccccc}
\toprule
Method         & Acc.           & Time          & Storage      & CS Mem.      & Train Mem. & Data Access \\
\midrule
Full           & $99.3 \pm 0.04$           & 100\%         & 100\%        & N/A            & 100\%           & 100\%        \\
Random         & $\textbf{99.2} \pm 0.05$  & \textbf{11\%} & 100\%        & \textbf{10\%}  & \textbf{10\%}   & $\sim$100\%  \\
GRAD-MATCH     & $99.0 \pm 0.07$           & 53\%          & 100\%        & 100\%          & \textbf{10\%}   & 100\%        \\
GLISTER        & $98.8 \pm 0.08$           & 27\%          & 100\%        & 100\%          & \textbf{10\%}   & 100\%        \\
DQ             & $97.8 \pm 0.25$           & 68\%          & 100\%        & 100\%          & \textbf{10\%}   & \textbf{10\%} \\
TDDS           & $99.0 \pm 0.18$           & 172\%         & 100\%        & 100\%          & \textbf{10\%}   & \textbf{10\%} \\
\midrule
dreaMLearning  & $99.1 \pm 0.07$           & \textbf{11\%} & \textbf{58\%}& \textbf{10\%}  & \textbf{10\%}   & $\sim$100\%  \\
\bottomrule
\end{tabular}
\end{table}

\begin{table}[h]
\centering
\caption{CIFAR-10}
\label{tab:cifar10}
\begin{tabular}{lccc}
\toprule
Method                 & Acc.           & Time           & Storage      \\
\midrule
Full                   & $95.2 \pm 0.16$           & 100\%          & 100\%        \\
Random                 & $\textbf{90.2} \pm 0.22$  & \textbf{11\%}  & 100\%        \\
GRAD-MATCH             & $89.2 \pm 0.30$           & 29\%           & 100\%        \\
GLISTER                & $88.7 \pm 0.59$           & 19\%           & 100\%        \\
DQ                     & $78.7 \pm 0.64$           & 20\%           & 100\%        \\
TDDS                   & $85.2 \pm 0.08$           & 104\%          & 100\%        \\
\midrule
\textbf{dreaMLearning} & $\textbf{90.2} \pm 0.34$  & \textbf{11\%}  & \textbf{80\%}\\
\bottomrule
\end{tabular}
\end{table}

\begin{table}[h]
\centering
\caption{CIFAR-100}
\label{tab:cifar100}
\begin{tabular}{lccc}
\toprule
Method                 & Acc.           & Time           & Storage      \\
\midrule
Full                   & $78.2 \pm 0.19$           & 100\%          & 100\%        \\
Random                 & $\textbf{69.7} \pm 0.26$  & \textbf{11\%}  & 100\%        \\
GRAD-MATCH             & $61.5 \pm 0.42$           & 21\%           & 100\%        \\
GLISTER                & $58.3 \pm 0.68$           & 19\%           & 100\%        \\
DQ                     & $36.6 \pm 0.71$           & 26\%           & 100\%        \\
TDDS                   & $51.9 \pm 0.12$           & 106\%          & 100\%        \\
\midrule
\textbf{dreaMLearning} & $\textbf{69.7} \pm 0.39$  & \textbf{11\%}  & \textbf{73\%}\\
\bottomrule
\end{tabular}

\end{table}

\section{Conclusion and future work}
\label{sec:conclusion}





We introduce dreaMLearning, a unified framework that seamlessly integrates entropy-based generalized deduplication (EntroGeDe) with machine learning, enabling efficient training directly on compressed data without decompression. 
Our approach leverages EntroGeDe to produce condensed, weighted samples for regression tasks on tabular data, while incorporating lightweight random sampling and frequency-domain transformations for classification on high-dimensional image datasets. 
Extensive evaluations on datasets such as California Housing, MNIST, CIFAR-10, and CIFAR-100 demonstrate substantial gains in training speed, memory efficiency, and storage requirements, all while maintaining competitive accuracy. 
These advancements position dreaMLearning as a promising solution for scalable and efficient learning, particularly in distributed, federated, and edge computing environments.
However, the current scope of dreaMLearning is limited to tabular and image data, leaving its performance on other modalities, such as time series, text, or graphs, unexplored. 
The effectiveness of EntroGeDe hinges data-specific redundancy patterns, which may be compromised by temporal dependencies in time series data or the sparse structures in textual data.
To address these constraints, future work will extend dreaMLearning to diverse data modalities, integrate adaptive compression strategies, and further investigate its applications in resource-constrained settings to enhance its robustness and versatility.

\clearpage
\appendix

\section*{Appendix}
\label{sec:appendix}

The appendix aims to provide additional details for dreaMLearning focusing on an expanded performance evaluation on the highlighted datasets and problems of the main paper, implications on other key problems (e.g., linear regression, logistic regression) looking at issues of complexity and overall performance, as well as providing a motivation behind the use of the EntroGeDe scheme. Some discussions rely on additional, suitable datasets for the specific problems considered.

\section{Classification with different data fractions}
\label{sec:appendix_experiments}

In the main paper, we reported classification experiment results utilizing a 10\% subset of the entire dataset, a standard proportion for baseline evaluations.
This selection was used due to recommended values from the works presenting the comparison schemes we utilized, thus providing a reasonable degree of fairness in our comparisons.
In the following, we present supplementary results derived from experiments conducted on 1\%, 5\%, and 20\% fractions of the dataset to highlight the effects of this selection on the accuracy and training time.

\begin{figure}[h]
    \centering
    \begin{subfigure}[b]{0.49\textwidth}
        \includegraphics[width=\textwidth]{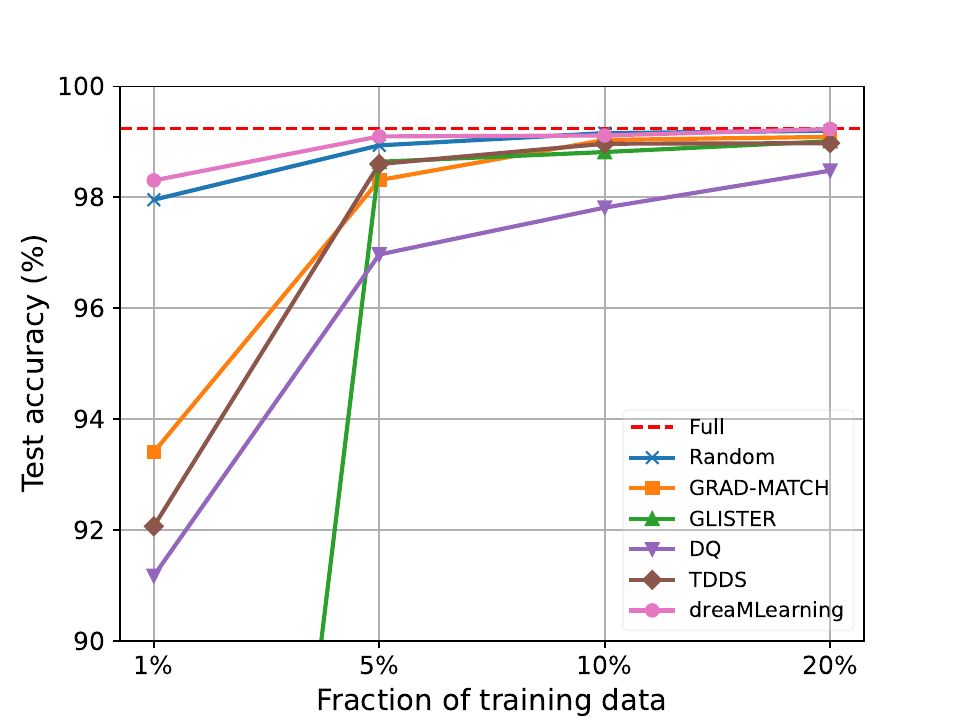}
        \caption{Accuracy vs. fraction of data}
        \label{fig:mnist_fraction_vs_acc}
    \end{subfigure}
    \hfill
    \begin{subfigure}[b]{0.49\textwidth}
        \includegraphics[width=\textwidth]{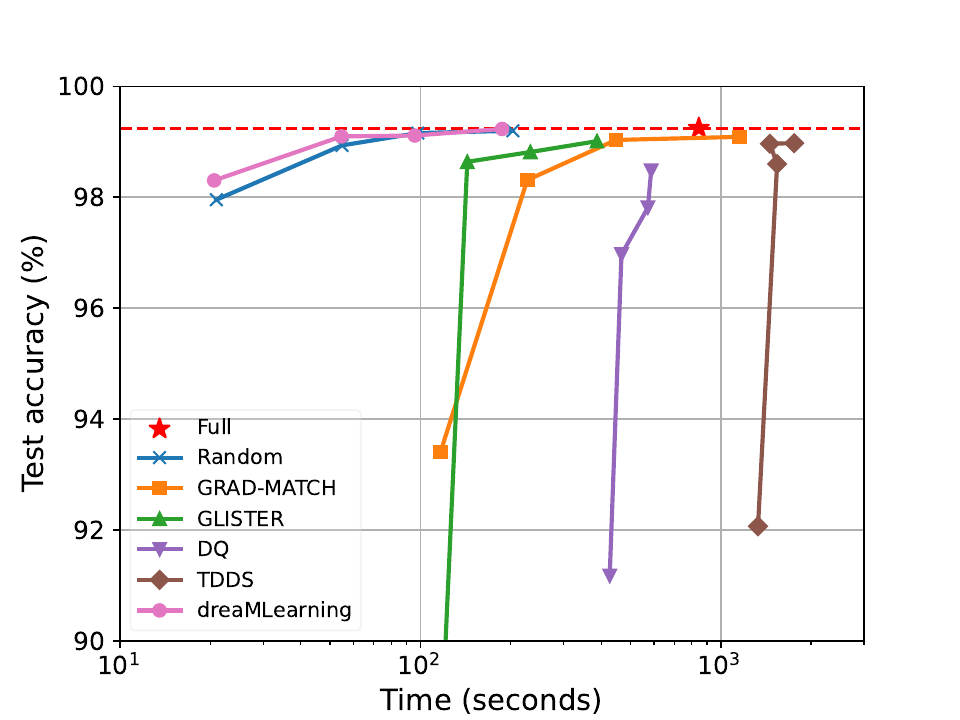}
        \caption{Accuracy vs. time}
        \label{fig:mnist_time_vs_acc}
    \end{subfigure}
    \caption{Training LeNet5 model on 1\%, 5\%, 10\% and 20\% of MNIST dataset.
    In (a), we show accuracy vs. fraction of data.
    In (b), we show accuracy vs. time.}
    \label{fig:mnist_experiments}
\end{figure}

\begin{figure}[h]
    \centering
    \begin{subfigure}[b]{0.49\textwidth}
        \includegraphics[width=\textwidth]{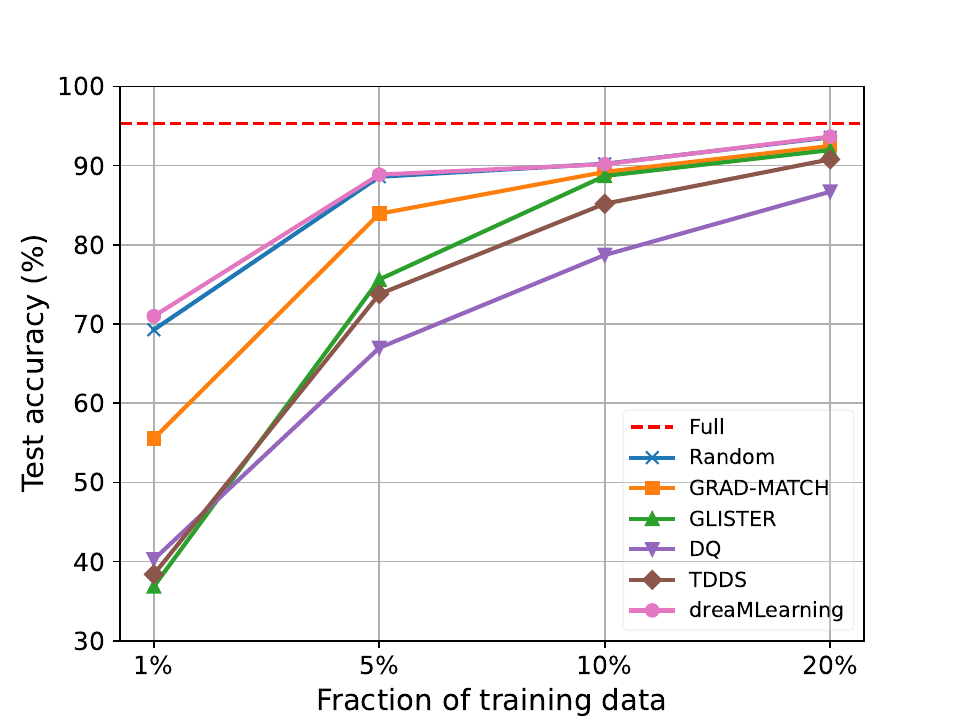}
        \caption{}
        \label{}
    \end{subfigure}
    \hfill
    \begin{subfigure}[b]{0.49\textwidth}
        \includegraphics[width=\textwidth]{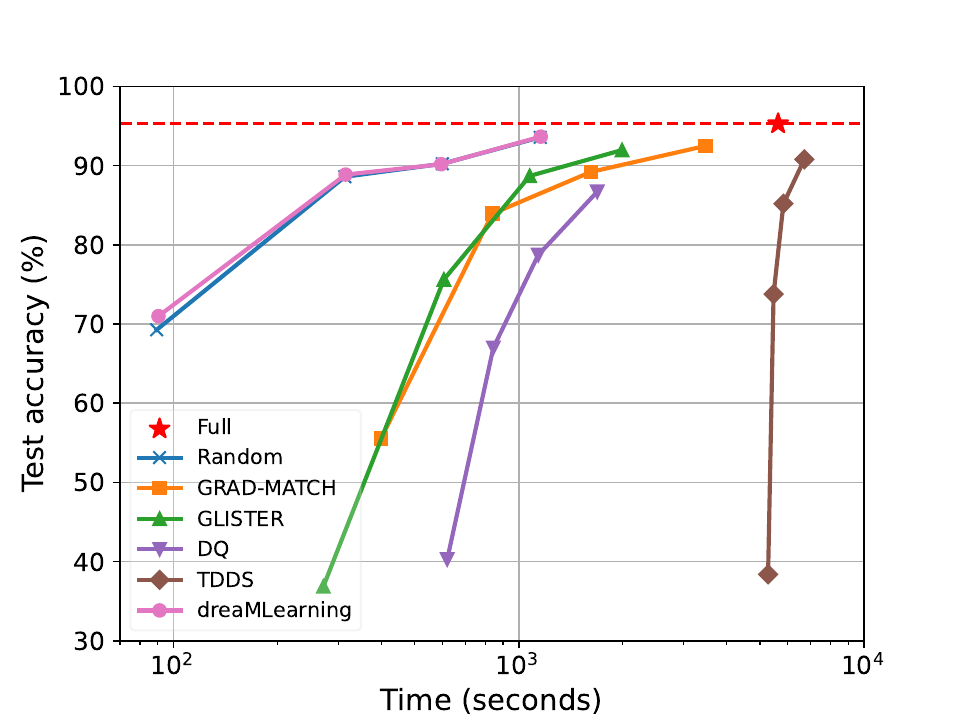}
        \caption{}
        \label{}
    \end{subfigure}
    \caption{Training ResNet18 model on 1\%, 5\%, 10\% and 20\% of CIFAR10 dataset.
    In (a), we show the fraction of data used vs. accuracy.
    In (b), we show the time vs. accuracy.}    
    \label{fig:cifar10_experiments}
\end{figure}

\begin{figure}[h]
    \centering
    \begin{subfigure}[b]{0.49\textwidth}
        \includegraphics[width=\textwidth]{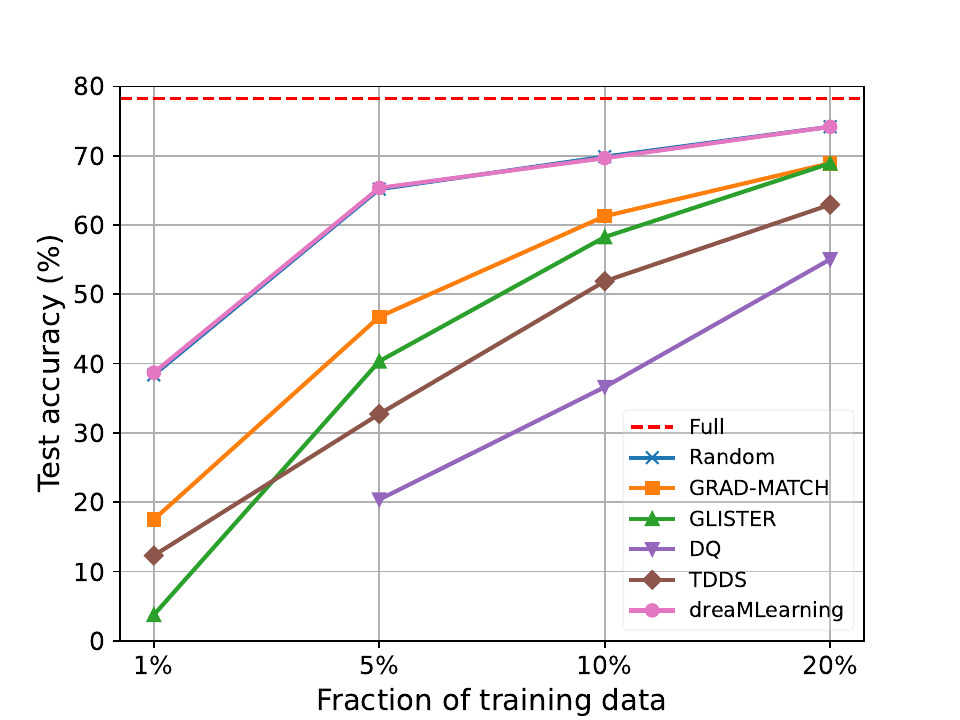}
        \caption{}
        \label{}
    \end{subfigure}
    \hfill
    \begin{subfigure}[b]{0.49\textwidth}
        \includegraphics[width=\textwidth]{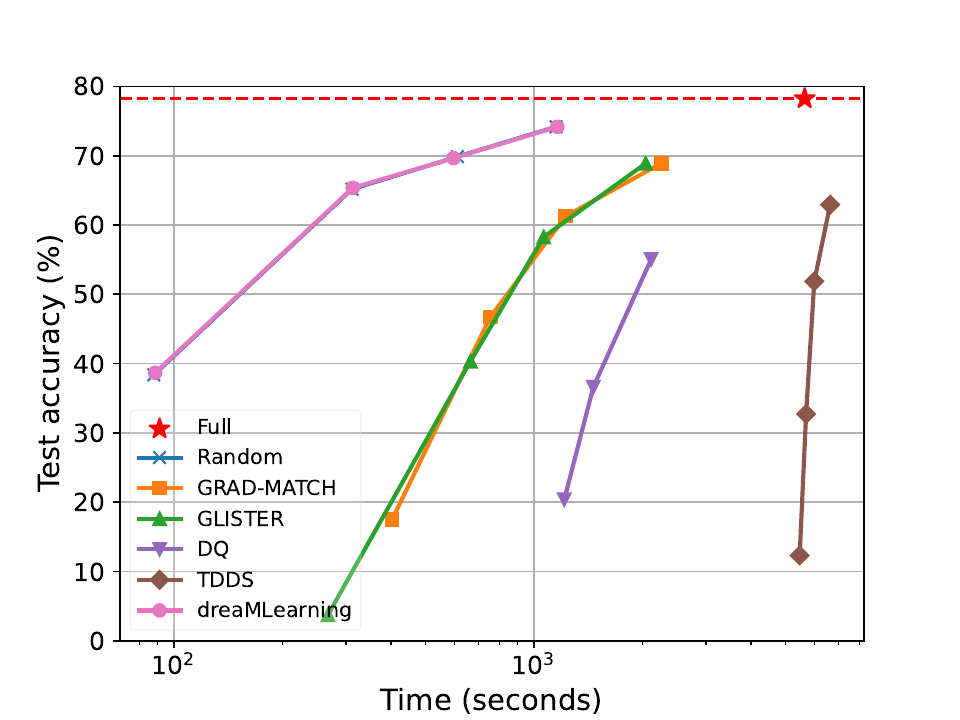}
        \caption{}
        \label{}
    \end{subfigure}
    \caption{Training ResNet18 model on 1\%, 5\%, 10\% and 20\% of CIFAR100 dataset.
    In (a), we show the fraction of data used vs. accuracy.
    In (b), we show the time vs. accuracy.}   
    \label{fig:cifar100_experiments}
\end{figure}

Figures~\ref{fig:mnist_experiments}, \ref{fig:cifar10_experiments} and \ref{fig:cifar100_experiments} show the results for MNIST, CIFAR10, and CIFAR100 datasets, respectively.
Results for 1\%, 5\%, and 20\% fractions are summarized in Tables~\ref{tab:mnist_results}, \ref{tab:cifar10_results}, and \ref{tab:cifar100_results} for convenience.
As in the paper, we focus on analyzing the performance of dreaMLearning and the selected baseline methods, including GRAD-MATCH, GLISTER, DQ, and TDDS.
DQ lacks results for a 1\% fraction of the CIFAR100 dataset, due to the fact that DQ divides the training data into 10 bins, each containing 5000 data points with 50 per class. Thus, selecting 1\% of 50 data points per class is infeasible, preventing this configuration. However, DQ is shown to have consistently lower performance in accuracy and training time compared to the other schemes (and particularly dreaMLearning) even when using higher fractions of data. For example, dreaMLearning has an accuracy of around 40\% at 1\% of the training data, while DQ has an accuracy of around 20\% using 5\% of the data.


\paragraph{Accuracy vs. fraction of data}
For all datasets and data fractions, the proposed dreaMLearning method consistently outperforms the baseline methods.
The performance gap between dreaMLearning and the baselines is especially pronounced when using smaller data fractions, such as 1\% and 5\%.
Furthermore, this gap becomes more significant for complex datasets like CIFAR10 and CIFAR100 compared to simpler datasets like MNIST.
For example, on the MNIST dataset, dreaMLearning achieves 98.3\% accuracy with 1\% of the data, while the best-performing baseline, GRAD-MATCH, achieves 93.4\% (while dreaMLearning is 5.57 times faster).

The accuracy gap increases in CIFAR10, where dreaMLearning achieves 71.0\% accuracy with 1\% of the data, compared to 55.5\% for GRAD-MATCH.
This performance difference further widens in CIFAR100 under the same data fraction of 1\%, namely, dreaMLearning achieves an accuracy of 38.7\% while GRAD-MATCH is only 17.5\%.
These results indicate that dreaMLearning is particularly effective at learning from smaller data fractions, especially in the context of more complex datasets.
 
 The performance gap narrows as the data fraction increases, although the is an indication of a \emph{sufficient} fraction of data required to start closing the gap. The more complex the dataset and problem, the more data is required for baseline schemes to close the gap to dreaMLearning. For example, at 5\% of data the gap is minimal for MNIST (99.1\% versus 98.6\% for GLISTER, which outperforms the other baseline schemes in this configuration) while the gap for CIFAR100 is considerably larger (65.4\% versus 46.8\%). It is also important to note that for MNIST at 5\% of data, dreaMLearning's accuracy is only 0.2 percentage points away from the performance achieved when training with all the data.
 Finally, note that baseline methods incur consistently higher computational costs for subset selection. 

\paragraph{Accuracy vs. time}
Let us consider a different scenario, namely how long time is required to achieve a similar accuracy allowing for other schemes to use a larger fraction of the data.
The time consists of both subset selection and training time on the selected subset.
Our results show that dreaMLearning achieves a specified accuracy level an order of magnitude faster than baseline methods.

On the MNIST dataset, dreaMLearning attains 98.3\% accuracy in 21 seconds.
In contrast, GLISTER requires approximately $7 \times$ longer to reach 98.6\%, GRAD-MATCH takes $11 \times$ longer for 98.3\%, DQ needs $28 \times$ longer for 98.5\%, and TDDS takes $75 \times$ longer for 98.6\%.
This efficiency stems from the effective and lightweight random sampling strategy of dreaMLearning, which bypasses costly subset selection by directly sampling compressed data.
Comparable trends appear in the CIFAR10 and CIFAR100 datasets.
As the data fraction increases, baseline methods exhibit substantially higher runtimes due to the growing computational cost of subset selection.
TDDS consistently incurs the longest runtime, even exceeding full dataset training, as it requires training on the entire dataset prior to subset selection.
These results highlight the substantial advantage of dreaMLearning in time efficiency.

\begin{table*}[h]
\centering
\caption{MNIST: Accuracy (\%) and Time (seconds) for different data fractions}
\label{tab:mnist_results}
\resizebox{\textwidth}{!}{
\begin{tabular}{lcccccc}
\toprule
\multirow{2}{*}{Method} 
    & \multicolumn{2}{c}{1\%} 
    & \multicolumn{2}{c}{5\%} 
    & \multicolumn{2}{c}{20\%} \\
 & Acc. & Time 
    & Acc. & Time 
    & Acc. & Time \\
\midrule
Full           & $99.3 \pm 0.04$      & $846 \pm 7.4$      & $99.3 \pm 0.04$      & $846 \pm 7.4$      & $99.3 \pm 0.04$      & $846 \pm 7.4$      \\
Random         & $98.0 \pm 0.44$      & $21 \pm 0.3$       & $98.9 \pm 0.30$      & $55 \pm 0.3$       & $99.2 \pm 0.09$      & $203 \pm 4.0$        \\
GRAD-MATCH     & $93.4 \pm 1.69$      & $117 \pm 1.9$      & $98.3 \pm 1.11$      & $227 \pm 1.0$      & $99.1 \pm 0.10$      & $1157 \pm 9.6$     \\
GLISTER        & $65.0 \pm 11.79$     & $75 \pm 1.7$       & $98.6 \pm 0.13$      & $143 \pm 0.3$      & $99.0 \pm 0.16$      & $387 \pm 2.2$      \\
DQ             & $91.2 \pm 1.94$      & $427 \pm 3.5$      & $97.0 \pm 0.18$      & $467 \pm 3.7$      & $98.5 \pm 0.09$      & $587 \pm 3.6$      \\
TDDS           & $92.1 \pm 1.35$      & $1331 \pm 17.4$    & $98.6 \pm 0.12$      & $1541 \pm 129.3$       & $99.2 \pm 0.09$      & $1757 \pm 27.7$        \\
\midrule
dreaMLearning  & $\textbf{98.3} \pm \textbf{0.10}$ & $\textbf{21} \pm \textbf{0.4}$ & $\textbf{99.1} \pm \textbf{0.11}$ & $\textbf{55} \pm \textbf{0.4}$ & $\textbf{99.2} \pm \textbf{0.08}$ & $\textbf{187} \pm \textbf{1.4}$ \\
\bottomrule
\end{tabular}
}
\end{table*}

\begin{table*}[h]
\centering
\caption{CIFAR10: Accuracy (\%) and Time (seconds) for different data fractions}
\label{tab:cifar10_results}
\resizebox{\textwidth}{!}{
\begin{tabular}{lcccccc}
\toprule
\multirow{2}{*}{Method} 
    & \multicolumn{2}{c}{1\%} 
    & \multicolumn{2}{c}{5\%} 
    & \multicolumn{2}{c}{20\%} \\
 & Acc. & Time 
    & Acc. & Time 
    & Acc. & Time \\
\midrule
Full           & $95.3 \pm 0.16$      & $5640 \pm 14.2$      & $95.3 \pm 0.16$      & $5640 \pm 14.2$      & $95.3 \pm 0.16$      & $5640 \pm 14.2$      \\
Random         & $69.3 \pm 1.71$      & $\textbf{89} \pm \textbf{0.4}$       & $88.6 \pm 0.42$      & $\textbf{314} \pm \textbf{0.3}$       & $93.6 \pm 0.12$      & $\textbf{1154} \pm \textbf{3.0}$        \\
GRAD-MATCH     & $55.5 \pm 1.52$      & $399 \pm 5.5$      & $83.9 \pm 0.38$      & $838 \pm 6.7$      & $92.5 \pm 0.15$      & $3468 \pm 12.2$     \\
GLISTER        & $36.9 \pm 2.34$     & $272 \pm 2.6$       & $75.6 \pm 0.77$      & $607 \pm 7.2$      & $92.0 \pm 0.10$      & $1991 \pm 93.7$      \\
DQ             & $40.3 \pm 1.03$      & $621 \pm 5.6$      & $67.0 \pm 1.63$      & $844 \pm 3.1$      & $86.7 \pm 0.58$      & $1688 \pm 4.3$      \\
TDDS           & $38.4 \pm 0.91$      & $5275 \pm 52.4$    & $73.8 \pm 0.66$      & $5471 \pm 24.3$       & $90.8 \pm 0.23$      & $6720 \pm 178.3$        \\
\midrule
dreaMLearning  & $\textbf{71.0} \pm \textbf{0.69}$ & $90 \pm 2.5$ & $\textbf{88.9} \pm \textbf{0.27}$ & $315 \pm 0.7$ & $\textbf{93.7} \pm \textbf{0.09}$ & $1158 \pm 2.7$ \\
\bottomrule
\end{tabular}
}
\end{table*}

\begin{table*}[h]
\centering
\caption{CIFAR100: Accuracy (\%) and Time (seconds) for different data fractions}
\label{tab:cifar100_results}
\resizebox{\textwidth}{!}{
\begin{tabular}{lcccccc}
\toprule
\multirow{2}{*}{Method} 
    & \multicolumn{2}{c}{1\%} 
    & \multicolumn{2}{c}{5\%} 
    & \multicolumn{2}{c}{20\%} \\
 & Acc. & Time 
    & Acc. & Time 
    & Acc. & Time \\
\midrule
Full           & $78.2 \pm 0.19$      & $5640 \pm 21.6$      & $78.2 \pm 0.19$      & $5640 \pm 21.6$      & $78.2 \pm 0.19$      & $5640 \pm 21.6$      \\
Random         & $38.4 \pm 0.62$      & $\textbf{88} \pm \textbf{0.2}$       & $65.2 \pm 0.32$      & $\textbf{312} \pm \textbf{0.7}$       & $74.2 \pm 0.26$      & $\textbf{1151} \pm \textbf{3.4}$        \\
GRAD-MATCH     & $17.5 \pm 0.25$      & $403 \pm 3.8$      & $46.8 \pm 0.75$      & $756 \pm 8.2$      & $68.9 \pm 0.28$      & $2254 \pm 39.8$     \\
GLISTER        & $3.7 \pm 1.35$     & $268 \pm 1.9$       & $40.3 \pm 0.51$      & $665 \pm 27.7$      & $68.9 \pm 0.36$      & $2041 \pm 74.1$      \\
DQ             & NA      & NA      & $20.4 \pm 0.64$      & $1212 \pm 11.3$      & $55.0 \pm 0.46$      & $2113 \pm 4.8$      \\
TDDS           & $12.3 \pm 0.33$      & $5462 \pm 28.4$    & $32.7 \pm 0.43$      & $5692 \pm 39.8$       & $62.9 \pm 0.22$      & $6640 \pm 19.0$        \\
\midrule
dreaMLearning  & $\textbf{38.7} \pm \textbf{0.43}$ & $89 \pm 0.2$ & $\textbf{65.4} \pm \textbf{0.27}$ & $313 \pm 0.5$ & $\textbf{74.2} \pm \textbf{0.22}$ & $1159 \pm 0.8$ \\
\bottomrule
\end{tabular}
}
\end{table*}

\section{The effectiveness of entropy based GeDe}
\label{sec:appendix_entropy}

\begin{figure}[htbp]
    \centering
    \begin{minipage}{0.48\textwidth}
        \centering
        \includegraphics[width=\linewidth]{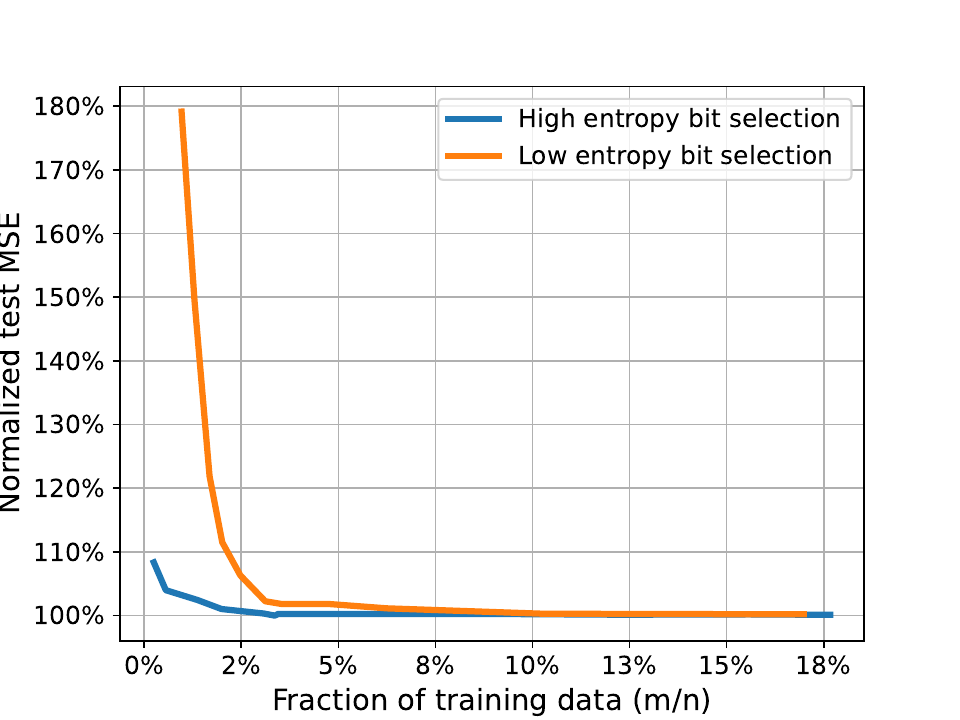}
        \caption{Compressed data fraction vs. MSE.}
        \label{fig:samples_vs_accuracy}
    \end{minipage}\hfill
    \begin{minipage}{0.48\textwidth}
        \centering
        \includegraphics[width=\linewidth]{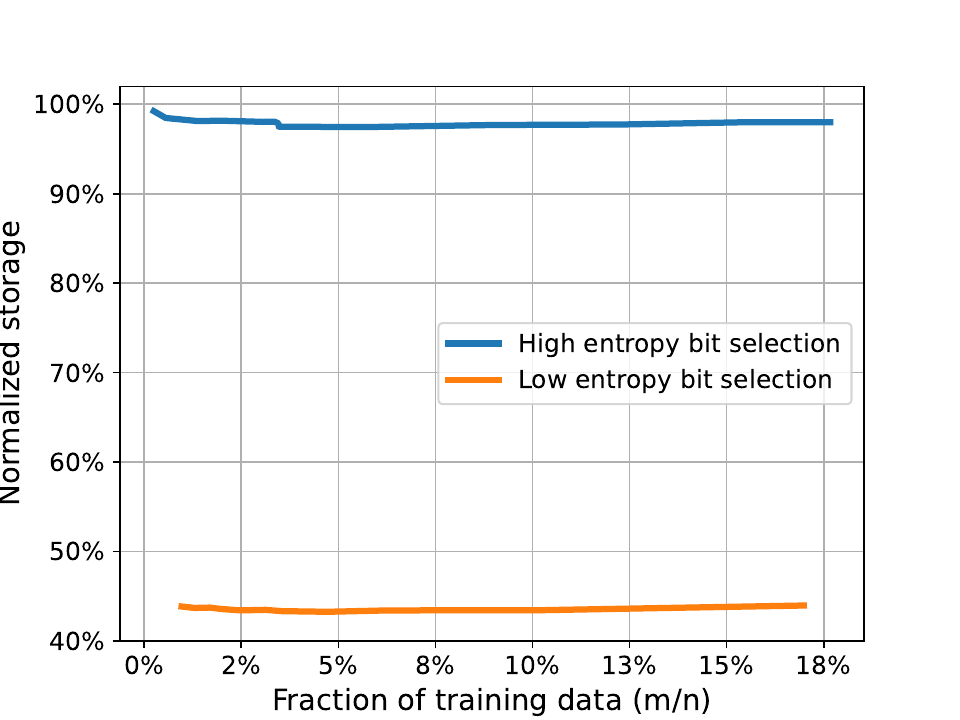}
        \caption{Compressed data fraction vs. storage.}
        \label{fig:samples_vs_compression_ratio}
    \end{minipage}
\end{figure}

In GeDe-based compression methods, selecting base bits is crucial for performance, as it determines how data is split into bases and deviations for compression.
The proposed EntroGeDe method selects base bits based on their entropy, which measures information content at each bit position.
The entropy is calculated as:
\begin{equation}
H(X) = - p_i \log_2 p_i - (1 - p_i) \log_2 (1 - p_i),
\end{equation}
where $p_i$ is the probability of a bit being 1 at the $i$-th position.
High-entropy bits, with balanced 0s and 1s, indicate high information content and are prioritized for analytics.
Conversely, low-entropy bits, with skewed distributions, contain more redundancy and are selected for compression.
EntroGeDe leverages high-entropy bits to generate the condensed samples, and low-entropy bits for effective compression (deduplication).
The approach is relatively simple compared to previous GeDe methods, e.g., GreedyGD, due to its ability to manage much larger dimensions in the data than previous schemes.

We validate effectiveness of EntroGeDe using the California Housing dataset~\cite{pace1997sparse}, comparing high- and low-entropy bit selections on linear regression MSE and storage.
Figure~\ref{fig:samples_vs_accuracy} shows that high-entropy bit selection significantly reduces MSE, confirming superior information retention.
Figure~\ref{fig:samples_vs_compression_ratio} demonstrates that low-entropy bit selection substantially lowers storage needs, validating its compression efficiency.
Therefore, EntroGeDe was designed to effectively combine high- and low-entropy bit selection to optimize both information retention and compression, providing an improved accuracy-compression trade-off inherent in existing GeDe methods.


\section{Complexity reduction analysis of dreaMLearning}

Training on compressed data substantially reduces complexity compared to using the full dataset, owing to fewer samples.
The complexity reduction of dreaMLearning for linear regression and classification tasks is analyzed below under various considerations.

\subsection{Linear regression tasks}

Consider a dataset of $n$ samples, where each sample is represented as a feature vector $\mathbf{x}_i \in \mathbb{R}^d$ and a target value $y_i \in \mathbb{R}$.
Linear regression estimates a parameter vector $\boldsymbol{\theta} \in \mathbb{R}^d$ to model the relationship between inputs and targets via $\hat{y}_i = \mathbf{x}_i^T \boldsymbol{\theta}$.
In dreaMLearning, we employ $m (m \ll n)$ compressed samples, denoted $\mathbf{x}_j^c$ and $y_j^c$.
These samples are generated by EntroGeDe, with associated weights $w_j$ reflecting the number of original samples condensed into each compressed sample.
The optimal $\boldsymbol{\theta}$ minimizes the error between targets and predicted values, typically using the mean squared error (MSE) loss function, defined as:

\begin{equation}
    J(\boldsymbol{\theta}) = \frac{1}{2n} \sum_{i=1}^{n} \left( \mathbf{x}_i^T \boldsymbol{\theta} - y_i \right)^2.
\end{equation}

\subsubsection{Gradient descent}
Gradient descent iteratively updates model parameters to minimize the MSE loss function.
The update rule is
\begin{equation}
\boldsymbol{\theta}_{t+1} = \boldsymbol{\theta}_{t} - \frac{\alpha}{n} \sum_{i=1}^{n} \left( \mathbf{x}_i^T \boldsymbol{\theta}_t - y_i \right) \mathbf{x}_i,
\end{equation}
where $\boldsymbol{\theta}_{t}$ is the parameter vector at iteration $t$, $\alpha$ is the learning rate.
This incurs $O(nd)$ time complexity per iteration, as gradients are computed for all $n$ samples, yielding $O(knd)$ for $k$ iterations.

In dreaMLearning, the update rule for compressed samples becomes
\begin{equation}
\boldsymbol{\theta}_{t+1}^c = \boldsymbol{\theta}_{t}^c - \frac{\alpha}{n} \sum_{j=1}^{m} w_j \left( {\mathbf{x}_j^c}^T \boldsymbol{\theta}_t - y_j^c \right) \mathbf{x}_j^c.
\end{equation}
This requires $O(md)$ time complexity per iteration, as gradients are computed for only $m$ samples.
Although the weight $w_j$ introduce minor overhead, the computational structure remains unchanged.
Thus, dreaMLearning reduces time complexity from $O(knd)$ to $O(kmd)$, a significant improvement when $m \ll n$.
Despite potentially more iterations, the smaller sample size substantially lowers overall time complexity.

\subsubsection{Optimal Solution}
Alternatively, linear regression has a closed form solution, which directly computes optimal parameters as
\begin{equation}
\begin{gathered}
\nabla_{\boldsymbol{\theta}} J(\boldsymbol{\theta}) = \boldsymbol{0} \\
\Rightarrow \frac{1}{2n} \nabla_{\boldsymbol{\theta}} \left\| \mathbf{X} \boldsymbol{\theta} - \mathbf{y} \right\|_2^2 = \boldsymbol{0} \\
\Rightarrow \mathbf{X}^\mathrm{T} \mathbf{X} \boldsymbol{\theta} - \mathbf{X}^\mathrm{T} \mathbf{y} = \boldsymbol{0} \\
\Rightarrow \boldsymbol{\theta}^* = \left( \mathbf{X}^{\mathrm{T}} \mathbf{X} \right)^{-1} \mathbf{X}^{\mathrm{T}} \mathbf{y}.
\end{gathered}
\end{equation}
This exact solution requires no hyperparameter tuning but is computationally intensive for large or high-dimensional datasets, where we consider $n$ to be the number of samples, and $d$ the dimensionality of the dataset.
The time complexity is driven by the matrix multiplication step $\mathbf{X}^{\mathrm{T}} \mathbf{X}$, which is $O(nd^2)$, and inversion step $\left( \mathbf{X}^{\mathrm{T}} \mathbf{X} \right)^{-1}$, which is $O(d^3)$. The time complexity is then $O(nd^2 + d^3)$.
Using dreaMLearning with $m \ll n$ compressed samples, the problem can be reduced to an approximate calculation of the form
\begin{equation}
\begin{gathered}
\nabla_{\boldsymbol{\theta}} J_c (\boldsymbol{\theta}) = \boldsymbol{0} \\
\Rightarrow \frac{1}{2n} \nabla_{\boldsymbol{\theta}} \left\| \mathbf{w}^{\frac{1}{2}} ( \mathbf{X}_{c} \boldsymbol{\theta} - \mathbf{y}_{c} ) \right\|_2^2 = \boldsymbol{0} \\
\Rightarrow \mathbf{X}_{c}^\mathrm{T} \mathbf{w} \mathbf{X}_{c} \boldsymbol{\theta} - \mathbf{X}_{c}^\mathrm{T} \mathbf{w} \mathbf{y}_{c} = \boldsymbol{0} \\
\Rightarrow \boldsymbol{\theta}^*_{c} = \left( \mathbf{X}_{c}^{\mathrm{T}} \mathbf{w} \mathbf{X}_{c} \right)^{-1} \mathbf{X}_{c}^{\mathrm{T}} \mathbf{w} \mathbf{y}_{c},
\end{gathered}
\end{equation}
where $\mathbf{w}$ is a diagonal matrix with weights $w_j$.
The additional complexity from weights is negligible as $\mathbf{w}$ is a $m \times m $ matrix.
The time complexity reduces to $O(md^2 + d^3)$, significantly lower when $m \ll n$.
This reduction enables dreaMLearning to make the calculation feasible for larger datasets, substantially saving time and computational resources.
The trade-off incurred is accuracy, as $\boldsymbol{\theta}^*_{c} \approx \boldsymbol{\theta}^*$.

Consider an example of California housing dataset shown in Figure~\ref{subfig:housing_n_samples_vs_mse}.
The original training set has $n=16,512$ samples and $d=8$ features.
Using dreaMLearning, we can reduce the dataset to $m=92$ samples, which is only 0.6\% of the original dataset.
The time complexity reduction by dreaMLearning is approximately  $165 \times$ for the optimal solution.
By gradient descent, the MSE from dreaMLearning is 4\% higher than that of the original dataset, and a similar difference is reasonably expected for the optimal solution.


\subsection{Classification tasks}

For classification tasks, computational complexity varies with model architecture.
Therefore, we consider the per-sample complexity of a model, denoted as $O(C)$, which is the cost of a forward and backward pass.
The per-epoch complexity is $O(nC)$, where $n$ is the number of training samples.
Using compressed data with $m$ samples ($m \ll n$), the per-epoch complexity reduces to $O(mC)$.
Consequently, total training complexity decreases from $O(EnC)$ to $O(E'mC)$, where $E$ and $E'$ are the epochs needed for convergence on the original and compressed datasets, respectively.
Typically, $E' \approx E$, reducing the computational cost by approximately $n/m$.
This significant reduction enhances training speed and lowers memory usage, making dreaMLearning highly efficient for large-scale classification with deep neural networks, as evidenced by our experimental results with ResNet18.

\section{Logistic regression} 
\begin{figure}[htbp]
    \centering
    \begin{subfigure}[b]{0.48\textwidth}
        \includegraphics[width=\textwidth]{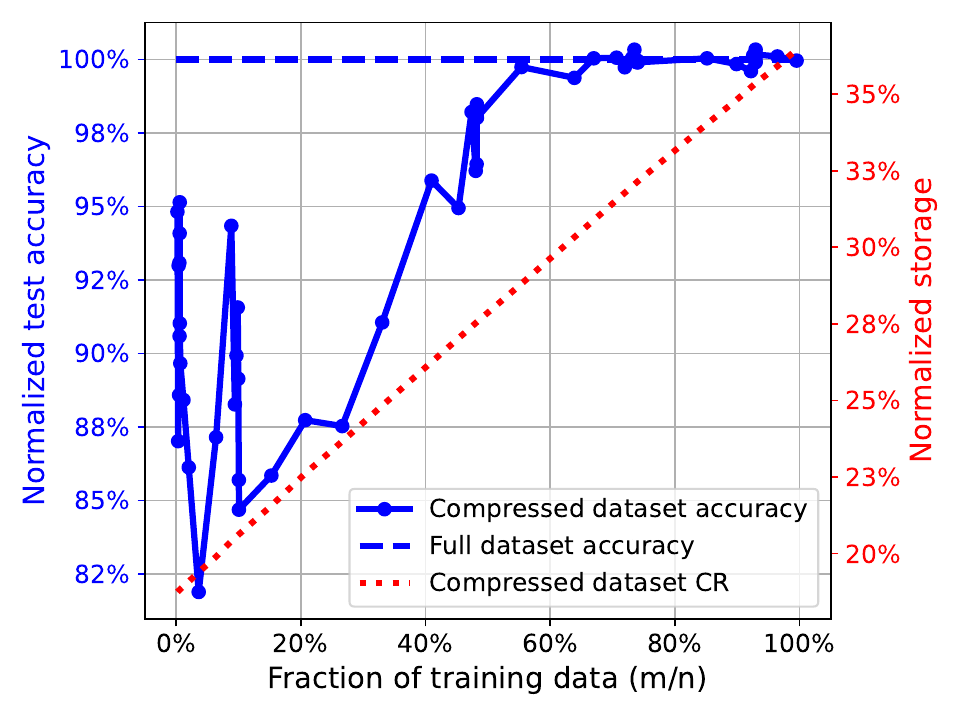}
        \caption{Default of credit card dataset.}
        \label{fig:credit_card}
    \end{subfigure}
    \hfill
    \begin{subfigure}[b]{0.48\textwidth}
        \includegraphics[width=\textwidth]{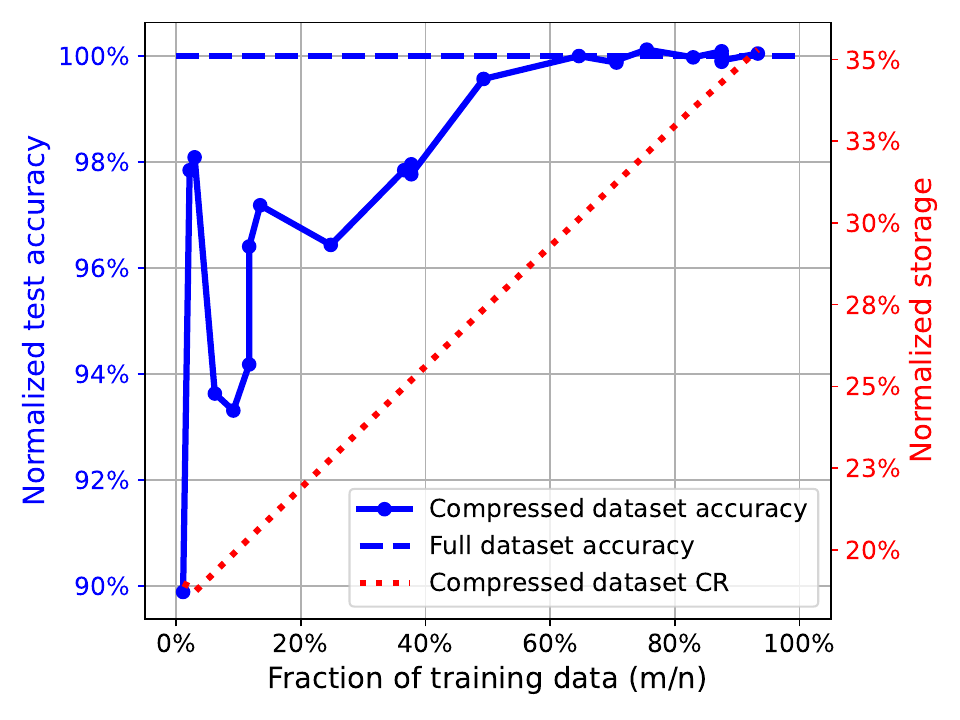}
        \caption{Ijcnn1 dataset.}
        \label{fig:ijcnn1}
    \end{subfigure}
    \caption{Comparison of full and compressed data for logistic regression tasks.}
    \label{fig:logistic_regression}
\end{figure}

We evaluate dreaMLearning for logistic regression on the Default of Credit Card~\cite{yeh2009comparisons} and IJCNN1 datasets~\cite{CC01a}.
The Default of Credit Card dataset comprises 30,000 samples with 23 features, split 80\% for training and 20\% for testing.
The IJCNN1 dataset includes 49,990 training and 91,701 test samples, each with 22 features.
Similar to linear regression tasks, we apply EntroGeDe to compress $n$ training data points to $m$ condensed ones with weights $w$.
The logistic regression model is trained using weighted gradient descent on these samples.
Figure~\ref{fig:logistic_regression} presents the results for both datasets considering the accuracy achieved (higher is better) and compression rate (lower is better). 
The test accuracy and storage requirement of compressed data are normalized relative to those of the full dataset.
dreaMLearning achieves performance comparable to full-dataset training with significantly fewer samples.
For the Default of Credit Card dataset, using 33\% of the data yields a 9\% accuracy loss with a 75\% storage footprint reduction (i.e., compression rate is roughly 0.25), while 55\% of the data achieves equivalent accuracy with a 70\% reduction.
For IJCNN1, 1\% of the data attains 90\% of full-training accuracy with an 80\% storage reduction, and 50\% of the data matches full-training accuracy with a 70\% reduction.
These results demonstrated that dreaMLearning enables substantial data and storage reduction for logistic regression, while maintaining accuracy close to that of full-data training.

\bibliographystyle{IEEEtran}
\bibliography{ref}

\begin{thebibliography}{10}
\providecommand{\url}[1]{#1}
\csname url@samestyle\endcsname
\providecommand{\newblock}{\relax}
\providecommand{\bibinfo}[2]{#2}
\providecommand{\BIBentrySTDinterwordspacing}{\spaceskip=0pt\relax}
\providecommand{\BIBentryALTinterwordstretchfactor}{4}
\providecommand{\BIBentryALTinterwordspacing}{\spaceskip=\fontdimen2\font plus
\BIBentryALTinterwordstretchfactor\fontdimen3\font minus \fontdimen4\font\relax}
\providecommand{\BIBforeignlanguage}[2]{{%
\expandafter\ifx\csname l@#1\endcsname\relax
\typeout{** WARNING: IEEEtran.bst: No hyphenation pattern has been}%
\typeout{** loaded for the language `#1'. Using the pattern for}%
\typeout{** the default language instead.}%
\else
\language=\csname l@#1\endcsname
\fi
#2}}
\providecommand{\BIBdecl}{\relax}
\BIBdecl

\bibitem{shen2024efficient}
L.~Shen, Y.~Sun, Z.~Yu, L.~Ding, X.~Tian, and D.~Tao, ``On efficient training of large-scale deep learning models,'' \emph{ACM Computing Surveys}, vol.~57, no.~3, pp. 1--36, 2024.

\bibitem{menghani2023efficient}
G.~Menghani, ``Efficient deep learning: A survey on making deep learning models smaller, faster, and better,'' \emph{ACM Computing Surveys}, vol.~55, no.~12, pp. 1--37, 2023.

\bibitem{zhou2022dataset}
Y.~Zhou, E.~Nezhadarya, and J.~Ba, ``Dataset distillation using neural feature regression,'' \emph{Advances in Neural Information Processing Systems}, vol.~35, pp. 9813--9827, 2022.

\bibitem{nguyen2021dataset}
T.~Nguyen, R.~Novak, L.~Xiao, and J.~Lee, ``Dataset distillation with infinitely wide convolutional networks,'' \emph{Advances in Neural Information Processing Systems}, vol.~34, pp. 5186--5198, 2021.

\bibitem{wang2018dataset}
T.~Wang, J.-Y. Zhu, A.~Torralba, and A.~A. Efros, ``Dataset distillation,'' \emph{arXiv preprint arXiv:1811.10959}, 2018.

\bibitem{cazenavette2022dataset}
G.~Cazenavette, T.~Wang, A.~Torralba, A.~A. Efros, and J.-Y. Zhu, ``Dataset distillation by matching training trajectories,'' in \emph{Proceedings of the IEEE/CVF Conference on Computer Vision and Pattern Recognition}, 2022, pp. 4750--4759.

\bibitem{sener2018active}
O.~Sener and S.~Savarese, ``Active learning for convolutional neural networks: A core-set approach,'' in \emph{International Conference on Learning Representations}, 2018.

\bibitem{sinha2020small}
S.~Sinha, H.~Zhang, A.~Goyal, Y.~Bengio, H.~Larochelle, and A.~Odena, ``Small-gan: Speeding up gan training using core-sets,'' in \emph{International Conference on Machine Learning}.\hskip 1em plus 0.5em minus 0.4em\relax PMLR, 2020, pp. 9005--9015.

\bibitem{colemanselection}
C.~Coleman, C.~Yeh, S.~Mussmann, B.~Mirzasoleiman, P.~Bailis, P.~Liang, J.~Leskovec, and M.~Zaharia, ``Selection via proxy: Efficient data selection for deep learning,'' in \emph{International Conference on Learning Representations}.

\bibitem{tonevaempirical}
M.~Toneva, A.~Sordoni, R.~T. des Combes, A.~Trischler, Y.~Bengio, and G.~J. Gordon, ``An empirical study of example forgetting during deep neural network learning,'' in \emph{International Conference on Learning Representations}, 2019.

\bibitem{paul2021deep}
M.~Paul, S.~Ganguli, and G.~K. Dziugaite, ``Deep learning on a data diet: Finding important examples early in training,'' \emph{Advances in neural information processing systems}, vol.~34, pp. 20\,596--20\,607, 2021.

\bibitem{mirzasoleiman2020coresets}
B.~Mirzasoleiman, J.~Bilmes, and J.~Leskovec, ``Coresets for data-efficient training of machine learning models,'' in \emph{International Conference on Machine Learning}.\hskip 1em plus 0.5em minus 0.4em\relax PMLR, 2020, pp. 6950--6960.

\bibitem{killamsetty2021grad}
K.~Killamsetty, S.~Durga, G.~Ramakrishnan, A.~De, and R.~Iyer, ``Grad-match: Gradient matching based data subset selection for efficient deep model training,'' in \emph{International Conference on Machine Learning}.\hskip 1em plus 0.5em minus 0.4em\relax PMLR, 2021, pp. 5464--5474.

\bibitem{killamsetty2021glister}
K.~Killamsetty, D.~Sivasubramanian, G.~Ramakrishnan, and R.~Iyer, ``Glister: Generalization based data subset selection for efficient and robust learning,'' in \emph{Proceedings of the AAAI Conference on Artificial Intelligence}, vol.~35, no.~9, 2021, pp. 8110--8118.

\bibitem{zhang2024spanning}
X.~Zhang, J.~Du, Y.~Li, W.~Xie, and J.~T. Zhou, ``Spanning training progress: Temporal dual-depth scoring (tdds) for enhanced dataset pruning,'' in \emph{Proceedings of the IEEE/CVF Conference on Computer Vision and Pattern Recognition}, 2024, pp. 26\,223--26\,232.

\bibitem{underwood2024understanding}
R.~Underwood, J.~C. Calhoun, S.~Di, and F.~Cappello, ``Understanding the effectiveness of lossy compression in machine learning training sets,'' \emph{arXiv preprint arXiv:2403.15953}, 2024.

\bibitem{zhao2020improving}
X.~Zhao, M.~Hosseinzadeh, N.~Hudson, H.~Khamfroush, and D.~E. Lucani, ``Improving the accuracy-latency trade-off of edge-cloud computation offloading for deep learning services,'' in \emph{2020 IEEE Globecom Workshops (GC Wkshps}.\hskip 1em plus 0.5em minus 0.4em\relax IEEE, 2020, pp. 1--6.

\bibitem{quinlan2002venti}
S.~Quinlan and S.~Dorward, ``Venti: A new approach to archival data storage,'' in \emph{Conference on file and storage technologies (FAST 02)}, 2002.

\bibitem{meyer2012study}
D.~T. Meyer and W.~J. Bolosky, ``A study of practical deduplication,'' \emph{ACM Transactions on Storage (ToS)}, vol.~7, no.~4, pp. 1--20, 2012.

\bibitem{Vestergaard_2019b}
R.~Vestergaard, Q.~Zhang, and D.~E. Lucani, ``Generalized deduplication: Bounds, convergence, and asymptotic properties,'' in \emph{{IEEE} Global Communications Conference ({GLOBECOM})}, 2019.

\bibitem{Vestergaard_2020}
R.~Vestergaard, D.~E. Lucani, and Q.~Zhang, ``A randomly accessible lossless compression scheme for time-series data,'' in \emph{{IEEE} {INFOCOM}}, 2020.

\bibitem{hurst2022glean}
A.~Hurst, D.~E. Lucani, I.~Assent, and Q.~Zhang, ``Glean: Generalized deduplication enabled approximate edge analytics,'' \emph{IEEE Internet of Things Journal}, vol.~10, no.~5, pp. 4006--4020, 2022.

\bibitem{hurst2024greedygd}
A.~Hurst, D.~E. Lucani, and Q.~Zhang, ``{GreedyGD}: Enhanced generalized deduplication for direct analytics in {IoT},'' \emph{IEEE Transactions on Industrial Informatics}, vol.~20, no.~4, pp. 6954--6962, 2024.

\bibitem{Taurone_2023}
F.~Taurone, D.~E. Lucani, M.~Fehér, and Q.~Zhang, ``Change a bit to save bytes: Compression for floating point time-series data,'' in \emph{IEEE International Conference on Communications}, 2023, pp. 3756--3761.

\bibitem{Rask_2024}
C.~D. Rask and D.~E. Lucani, ``Rage for the machine: Image compression with low-cost random access for embedded applications,'' in \emph{IEEE International Conference on Image Processing (ICIP)}, 2024, pp. 1987--1993.

\bibitem{Hurst_2021}
A.~Hurst, Q.~Zhang, D.~E. Lucani, and I.~Assent, ``Direct analytics of generalized deduplication compressed {IoT} data,'' in \emph{{IEEE} Global Communications Conference ({GLOBECOM})}, 2021.

\bibitem{Taurone_2024}
F.~Taurone, J.~Dorsch, D.~Lucani, and Q.~Zhang, ``{triaGeD}: using compression for anomaly detection,'' in \emph{Data Compression Conference (DCC)}, 2024, pp. 588--588.

\bibitem{zhou2023dataset}
D.~Zhou, K.~Wang, J.~Gu, X.~Peng, D.~Lian, Y.~Zhang, Y.~You, and J.~Feng, ``Dataset quantization,'' in \emph{Proceedings of the IEEE/CVF International Conference on Computer Vision}, 2023, pp. 17\,205--17\,216.

\bibitem{iyer2021submodular}
R.~Iyer, N.~Khargoankar, J.~Bilmes, and H.~Asanani, ``Submodular combinatorial information measures with applications in machine learning,'' in \emph{Algorithmic Learning Theory}.\hskip 1em plus 0.5em minus 0.4em\relax PMLR, 2021, pp. 722--754.

\bibitem{pace1997sparse}
R.~K. Pace and R.~Barry, ``Sparse spatial autoregressions,'' \emph{Statistics \& Probability Letters}, vol.~33, no.~3, pp. 291--297, 1997.

\bibitem{lecun1998gradient}
Y.~LeCun, L.~Bottou, Y.~Bengio, and P.~Haffner, ``Gradient-based learning applied to document recognition,'' \emph{Proceedings of the IEEE}, vol.~86, no.~11, pp. 2278--2324, 1998.

\bibitem{krizhevsky2009learning}
A.~Krizhevsky, G.~Hinton \emph{et~al.}, ``Learning multiple layers of features from tiny images,'' 2009.

\bibitem{yeh2009comparisons}
I.-C. Yeh and C.-h. Lien, ``The comparisons of data mining techniques for the predictive accuracy of probability of default of credit card clients,'' \emph{Expert systems with applications}, vol.~36, no.~2, pp. 2473--2480, 2009.

\bibitem{CC01a}
C.-C. Chang and C.-J. Lin, ``{LIBSVM}: A library for support vector machines,'' \emph{ACM Transactions on Intelligent Systems and Technology}, vol.~2, pp. 27:1--27:27, 2011, software available at \url{http://www.csie.ntu.edu.tw/~cjlin/libsvm}.

\end{thebibliography}

\end{document}